# A Knapsack by Any Other Name: Presentation impacts LLM performance on NP-hard problems

**Alex Duchnowski**
Saarland University
aduchnowski@coli.uni-saarland.de

**Ellie Pavlick**
Brown University
ellie_pavlick@brown.edu

**Alexander Koller**
Saarland University
koller@coli.uni-saarland.de

## Abstract

To investigate the effect of problem presentation on LLMs' ability to solve optimization problems, we introduce the dataset of Everyday Hard Optimization Problems (EHOP), a collection of NP-hard problems expressed in natural language. EHOP includes problem formulations that could be found in computer science textbooks (e.g., graph coloring), versions that are dressed up as problems that could arise in real life (e.g., party planning), and variants with inverted rules. We find that state-of-the-art LLMs, across multiple prompting strategies, systematically solve textbook problems more accurately than their real-life and inverted counterparts. While reasoning models are more capable, they nonetheless show high variance across problem presentations, suggesting they lack a truly robust reasoning mechanism. We argue that this constitutes evidence that LLMs are still heavily dependent on what was seen in training and struggle to generalize to novel problems.

## 1 Introduction

Many real-world tasks that people face in their personal and professional lives are NP-hard optimization problems. Such problems are as diverse as planning family vacations, scheduling airline crews (Gopalakrishnan and Johnson, 2005), and allocating organ donations (Abraham et al., 2007). People rarely enjoy solving these problems, and they aren't particularly good at solving them either (Hidalgo-Herrero et al., 2013).

One of the most exciting promises of large language models (LLMs) is that they can help non-experts solve their real-world computational problems when they express them in natural language (NL). The hope is that a wide range of users across a wide range of tasks will be able to describe their problem to an LLM, and the LLM will handle the difficult task of "problem solving," i.e., recognizing that the real-world problem can be described in terms of a known computational problem and then solving that problem efficiently and optimally. In the case of NP-hard problems, this could potentially be accomplished either by the LLM solving the problem by itself, e.g., through chain-of-thought (CoT) reasoning (Fan et al., 2024), or by the LLM converting the NL description into a linear program (LP) to be solved with specialized tools (AhmadiTeshnizi et al., 2024).

| |
|---|
| **Textbook:** Given the undirected graph $G$, color its nodes such that no two adjacent nodes have the same color. Use as few colors as possible. |
| **Costumed (💔 Parties with Exes):** Your birthday is coming up, and you want to celebrate with all your friends. You do not want people who used to be in a relationship at the same party. How many parties do you need, and who should be invited to which party? |
| **Inverted:** Given the undirected graph $G$, color its nodes such that no two *non-adjacent* nodes have the same color. Use as few colors as possible. |

Figure 1: Variants of GRAPH COLORING in EHOP.

However, recent work has raised the question of "reasoning vs. reciting": are LLMs actually carrying out systematic problem-solving, or are they simply adapting solutions for similar problems in their training data (Mirzadeh et al., 2024; Wu et al., 2024)? LLMs that can only solve problems whose solution paths are documented on the Internet will not fulfill the promise of opening robust, general problem-solving to lay users.

In this paper, we contribute to the reasoning vs. reciting debate by introducing the dataset of Everyday Hard Optimization Problems (EHOP), which consists of NP-hard optimization problems presented in both textbook and real-world variants (see Figure 1 for an example). If LLMs perform reasoning, they should solve both variants at similar levels of accuracy. If they recite, we would expect textbook problems, for which solution strategies are presented explicitly on the Internet, to be easier. To enable this direct comparison, EHOP introduces "costumes" for three well-studied problems (GRAPH COLORING, KNAPSACK, and TRAVELING SALESMAN) that represent real-world situations with the same mathematical constraints. Furthermore, we add *inverted* variants of all problems, which fundamentally distort the solutions of the problems with a small change in problem formulation.

For standard LLMs, including GPT-4o (OpenAI, 2024), Llama 3.1 (Grattafiori et al., 2024), and Qwen3-32B (Yang et al., 2025) we find that the proportion of textbook problems solved optimally is substantially higher than for the inverted and costumed variants, often by more than 20 percentage points. This holds across all three base problems, for different degrees of problem instance difficulty, and across multiple prompting strategies. The best-performing approach uses LLMs to convert problems into LPs and solves the LPs with a separate tool, but the vulnerability to inversion and costuming persists.

The only models we tested that do not exhibit systematic performance degradations on our problem variants are DeepSeek-R1 (DeepSeek-AI, 2025), which is specifically trained for reasoning and Qwen3-32B in thinking mode. However, even these models display major fluctuations across variants and can underperform on costumes by over 20 percentage points in certain cases. Thus, even on state-of-the-art reasoning models, the presentation of the problem (textbook vs. costumed or inverted) greatly affects LLM performance, suggesting that general, "reasoning"-style problem solving with LLMs remains an open challenge.

## 2 Related Work

LLMs have been shown to perform remarkably well on benchmarks for complex problem-solving tasks, such as tool use (Yao et al., 2023), complex gameplay (Wang et al., 2023), and planning (Stein et al., 2024). This has been attributed to the ability of iterative prompting strategies such as CoT (Kojima et al., 2022; Wei et al., 2022) to perform general reasoning and problem solving.

However, recent work has raised the question of whether LLMs actually perform systematic reasoning, or whether they are "reciting" solution paths from their training data and adapting them gracefully to the inference-time problem (Wu et al., 2024; Kambhampati, 2024). The fact that LLM reasoners often degrade in accuracy for larger problem instances is one piece of evidence for the recitation hypothesis. Furthermore, as long as chains of thought are limited to a polynomial number of steps, transformers provably solve exactly the problems that can be solved in polynomial time (Merrill and Sabharwal, 2024), failing to cover most reasoning problems, for which no optimal polynomial algorithms are known.

In this paper, we focus on NP-hard optimization problems, with particular attention to the difference between textbook and everyday problems. Previous work has investigated the ability of LLMs to solve NP-hard optimization problems (e.g. Yang et al., 2024; Guo et al., 2024; Wu et al., 2025). Here we do not aim to further improve LLM-based optimization as such; our focus is on the impact of problem presentation on LLM performance. Nonetheless, we include OPRO (Yang et al., 2024), one of the leading LLM-based optimizers, as a prompting strategy in the evaluation.

Finally, there are a number of existing datasets for evaluating models on NP-hard problems. NPHardEval (Fan et al., 2024) looks only at textbook problems, including the three base problems we consider here. GraphArena (Tang et al., 2024) evaluates LLMs on NP-hard graph problems with a variety of large real-world graphs, and is also limited to textbook problems. NL4Opt (Ramamonjison et al., 2022) and NLP4LP (AhmadiTeshnizi et al., 2024) provide evaluation datasets on real-world NP-hard problems, but they are not linked to the underlying textbook problems. EHOP differs from all these datasets in that we present the exact same instances of the base problem both in textbook and real-world variants, making it possible for the first time to measure the impact of this distinction.

## 3 Everyday optimization problems

An optimization problem is called *NP-hard* if every problem that can be solved in non-deterministic polynomial time can be reduced to the problem in polynomial time (Garey and Johnson, 1979). While it is generally assumed that deterministic algorithms that solve NP-hard problems must have worst-case exponential runtime, problems in NP are still of lower computational complexity than, e.g., planning or reasoning. In

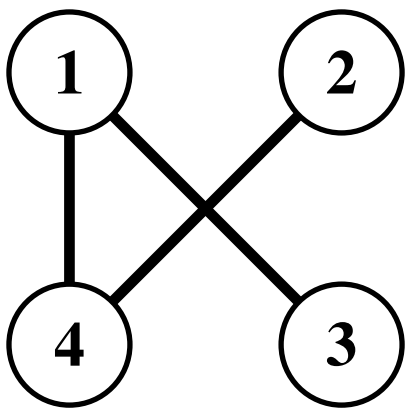

I have a network of 4 nodes, numbered 1 to 4, with various nodes being connected to one another. I want to color the nodes such that no two connected nodes have the same color.
The connections are as follows:
Node 1 and node 3 are connected.
Node 1 and node 4 are connected.
Node 2 and node 4 are connected.
How can I color the nodes using the fewest colors possible?

I am a teacher, and I want to assign my 4 students to different groups. I need the groups to focus, so I need to make sure that no two students who are friends with one another are in the same group, otherwise they may get distracted. I don't need the groups to all be the same size, but I want to minimize the total number of groups.
The friendships are as follows:
Student 1 and student 3 are friends.
Student 1 and student 4 are friends.
Student 2 and student 4 are friends.
Which group should each student be assigned to?

Figure 2: An example of a Graph Coloring problem instance with (truncated) Textbook and Student Groups presentations of the instance.

this paper, we focus on three well-known NP-hard optimization problems (which we refer to as *base problems*): Graph Coloring, Knapsack, and Traveling Salesman.

To construct the dataset of Everyday Hard Optimization Problems (EHOP), we first generate a number of random instances for each of the three base problems. Instances are concrete examples of a problem; for example, an instance of the Graph Coloring problem consists of a specific graph $G$ (see Figure 2). We present each instance in its Textbook form, which uses terminology typical for the problem; in addition, we dress it up in three real-world *costumes* and *invert* it. This yields a total of eight *variants* of each instance. Appendix C shows examples of all variants.

Not all instances of an NP-hard problem are equally difficult. We therefore ensure that experimental results remain comparable across variants, especially when we invert the problems.

### 3.1 Graph Coloring

An instance of the Graph Coloring problem consists of an undirected graph $G = (V, E)$. The task is to assign each node a color such that no two adjacent nodes have the same color, while using the fewest colors possible.

Inverted Graph Coloring asks for color assignments in which no two *non-adjacent* nodes have the same color.[1] For each instance $G$ of the base problem, we take the complement of $G$ as an instance of the inverted problem; it has an edge between two nodes if and only if there is no edge between them in $G$. Thus, the same coloring will solve the inverted problem on the inverted instance, ensuring identical difficulty.

[1]This is equivalent to the clique cover problem.

In addition to the **Textbook** variant, we have constructed three costumes that are not overtly about graph coloring:

**Student Groups.** $V$ represents a set of students, and $E$ represents friendships. A teacher wants to assign students to as few groups as possible, while ensuring that no student is distracted by a groupmate who is also a friend.

**Parties with Exes.** $V$ represents a person's set of friends, and $E$ represents which friends used to be in a romantic relationship with each other. This person wants to celebrate their birthday with their friends while avoiding awkwardness arising from exes being at the same party. They want to minimize the number of parties they have to plan.

**Taekwondo Tournament.** $V$ represents participants in a Taekwondo tournament, and $E$ represents which participants will be fighting one another in the tournament. The tournament organizer wants to assign participants to warm-up rooms without giving opponents the chance to study each other in advance of the competition.

### 3.2 Knapsack

An instance of the Knapsack problem consists of a knapsack with some capacity $C \in \mathbb{N}$ and a set of items with weights $w_1, \ldots, w_n \in \mathbb{N}$ and values $v_1, \ldots, v_n \in \mathbb{N}$. The task is to find a subset of items that maximizes the sum of the values of these items, under the constraint that their total weight must not be greater than $C$.

In inverted Knapsack, the task is to *minimize* the selected items' total value, with the constraint that the items' total weight must be *at least* $C$. For each instance of the base problem, we construct an instance of the inverted problem by setting the knapsack capacity to $\sum w_i - C$. Thus the optimal solution of the inverted instance consists of exactly the items that were *left out* of the knapsack in the original instance, ensuring equal difficulty.

We have constructed the following costumes:

**Lemonade Stand.** We have $C$ liters of lemonade to sell at our lemonade stand and would like to sell it for as much money as possible. Each of our $n$ customers offers to pay a price $v_i$ for $w_i$ liters of lemonade.

**Sightseeing.** We have $C$ hours to spend in Paris and would like to visit attractions that give us maximal total satisfaction. Each of the $n$ possible attractions will give us some satisfaction $v_i$ and take some time $w_i$ to visit.

**Party Planning.** We have a decoration budget $C$ for the party we are planning, and we wish to maximize the total coolness of our party. Each potential decoration item has a coolness score of $v_i$ and a price of $w_i$.

### 3.3 Traveling Salesman

An instance of the Traveling Salesman problem consists of a set $C = \{1, ..., n\}$ of cities, and for any pair of cities, we have a distance $d(i, j) \in \mathbb{N}$. The task is to find the shortest round trip that visits all the cities. That is, we are looking for a permutation $\pi : C \to C$ that minimizes

$$d(\pi_n, \pi_1) + \sum_{i=1}^{n-1} d(\pi_i, \pi_{i+1}).$$

Inverted Traveling Salesman changes the goal to *maximizing* the sum of the distances rather than minimizing it. For each instance of the base problem, we construct an instance of the inverted problem by converting each distance $d(i, j)$ to $m - d(i, j) + s$, where $m = \max d(i, j)$. We sample a random shift $s \in \{1, ..., n\}$ for each instance to maintain some variety of edge weights. This ensures that the optimal solutions of an instance and its inverted counterpart are the same.

We have constructed the following costumes:

**Task Schedule.** $C$ represents a set of tasks that have to be done daily, and $d$ represents the time it takes to modify one's workspace to transition between tasks. Note that the transition from one day to the next captures the term $d(\pi_n, \pi_1)$.

**Exercise Schedule.** As their New Year's resolution, a person will do a physical activity from a set $C$ every day, never repeating until they've exhausted the set, after which they will go through it again in the same order as before. They want to maximize the day-to-day variety of their activities by minimizing the similarity score $d$ between adjacent activities.

**UN Seating.** A staff member at the United Nations needs to figure out how to seat the representatives $C$ from various countries around a circular table. They want to minimize the total political tension $d$ between adjacent representatives.

## 4 Experiments

We use EHOP to measure the extent to which LLMs are vulnerable to changes in presentation when solving optimization problems.

### 4.1 Dataset

The EHOP dataset consists of two parts: EHOP-RANDOM and EHOP-HARD. Each of these two sub-datasets consists of 150 distinct instances of each of the three base problems (100 for Graph Coloring in EHOP-HARD, see below), presented in each of the eight variants (Textbook and three costumes × standard/inverted). In total, EHOP has 6800 natural language task descriptions. These task descriptions are designed to ensure that the prompts for the same instance are of similar length across variants (see Appendix C).

To create EHOP-RANDOM, we randomly generated 25 instances of each base problem for six different sizes: for Graph Coloring and Traveling Salesman, we generated instances with 4, 5, 6, 7, 8, and 9 nodes/cities, and for Knapsack, we generated instances with 4, 8, 12, 16, 20, and 24 items. These scales were chosen to represent a spectrum of difficulties ranging from easy to hard. We determined optimal solutions for each instance with an optimal solver.[2] EHOP-HARD contains instances of similar sizes, but ensures that all instances are hard to solve (more details in Section 5.3). We have released our code to enable the generation of more instances.[3]

### 4.2 Models and Prompting

We evaluate GPT-4o, Llama-3.1-70B Instruct, DeepSeek-R1, and Qwen3-32B on EHOP (see

[2]Solvers for Graph Coloring and Traveling Salesman were coded using the `gurobipy` package (Gurobi Optimization LLC, 2024), and Knapsack instances were solved using Google OR-Tools.

[3]The EHOP dataset and accompanying code are available at https://github.com/coli-saar/ehop. It is also worth noting that instances are generated without the use of any LLMs.

Appendix A for model details). For each LLM, we evaluate a number of prompting strategies; the detailed prompts are in Appendix D. The **One-Shot** strategy prompts the LLM for a solution to the NL task description, with a single example and its optimal answer prepended to the prompt. In the **Zero-Shot CoT** strategy, the task description is followed by the sentence "Let's think step by step" (Kojima et al., 2022). The **One-Shot CoT** strategy presents the same example used in the one-shot case, this time with an answer text that includes a chain of thought resulting in a solution (Wei et al., 2022).

We also implemented an **ILP Python** prompting strategy, which prompts the LLM to translate the problem instance into Python code[4] that calls the Gurobi solver on an Integer Linear Program (ILP) encoding of the instance (Gurobi Optimization LLC, 2024), cf. AhmadiTeshnizi et al. (2024). Thus, ILP Python does not attempt to solve the problem through LLM reasoning; the problem is solved exactly and optimally by Gurobi, and the LLM merely translates the NL specifications to code and then translates the code's output back into NL. If the code generated by the LLM produces an error, we halt the process and count it as a failure.

We also include a limited evaluation on **OPRO** (Yang et al., 2024), a leading approach for solving optimization problems with LLMs through repeated prompting. Due to the very high inference cost of OPRO (up to 80 LLM calls per instance), we evaluate it only on the instances of the largest and the second-smallest size of each variant; the smallest instances do not admit enough solutions to run OPRO effectively. All OPRO experiments are performed with GPT-4o.

Finally, we compare LLM performance on each problem to **greedy baselines**. For GRAPH COLORING, the greedy heuristic colors each node with the smallest color (where colors are represented by the numbers 1, 2, ...) that does not conflict with any neighbors that are already colored. Nodes are colored in descending order of number of neighbors. For KNAPSACK, the strategy goes through the items in descending order of density (value divided by weight), adding each item to the knapsack if it fits in the remaining capacity. For TRAVELING SALESMAN, we use the strategy of always moving to the closest unvisited city. We apply the greedy baselines directly to the original problem instances. These greedy strategies are linear-time algorithms which always produce valid solutions but give no guarantee of optimality.

### 4.3 Evaluation

We run all non-reasoning models with all prompting strategies and all reasoning models with zero-shot and ILP Python strategies for all instances in EHOP.

We classify the correctness of the outputs using the following scheme. An **incompatible** response is syntactically flawed; it can't be parsed as a solution to the problem. An **erroneous** response can be parsed as a solution, but it violates constraints of the underlying problem; for example, it assigns adjacent nodes in GRAPH COLORING the same color. Among the remaining responses, we distinguish between **optimal** and **suboptimal** solutions, depending on whether they find a configuration that optimizes the objective as much as possible. ILP Python can additionally produce **ILP code failures** if the LLM-generated code cannot be executed without errors. See Appendix B for examples of each result category.

Our main evaluation metric is **optimization accuracy:** the proportion of instances that were solved optimally. We do not evaluate the distances of the generated solutions from the optima, as such a measure only makes sense if all solutions are optimal or suboptimal. In our experiments, a varying proportion of solutions is incompatible or erroneous, distorting such a metric.

## 5 Results

### 5.1 Scaling to larger instances is hard

Figure 3 gives an overview of the optimization accuracy for each textbook problem, as a function of input size. One-Shot and Zero-Shot CoT are not shown in the plot to enhance readability; they perform worse than the other three (see Appendix E). We find that accuracy drops as instances are scaled up. This trend continues for larger sizes than those in EHOP; even ILP-Python with GPT-4o is below 5% optimization accuracy on textbook GRAPH COLORING instances with 12 and 15 nodes.

[4] We chose Python as the ILP specification syntax because this has been shown to outperform LLM translations into domain-specific languages (Bogin et al., 2024). We include results for specifying the ILPs in the domain-specific LP file format in Appendix E.

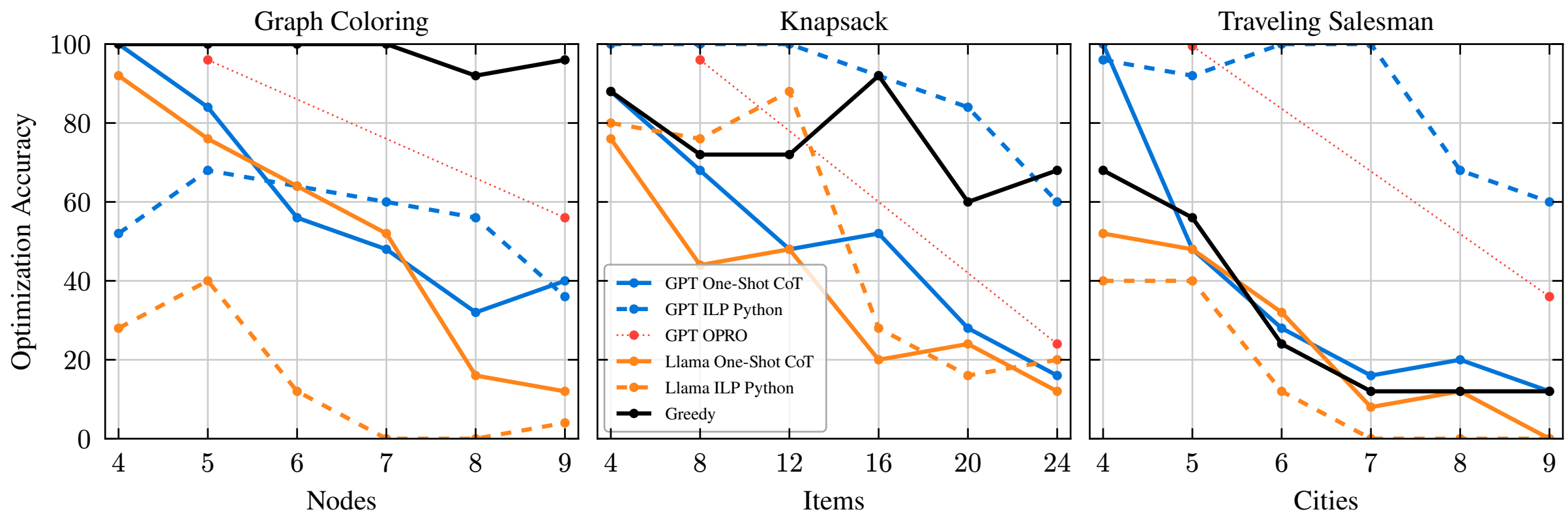

Figure 3: Optimization accuracy of GPT and Llama on the textbook variants in EHOP-RANDOM as a function of instance size. See Appendix F for Qwen results, which follow a similar pattern and are excluded here for readability.

| Problem | Variant | One-Shot | | | Zero-Shot CoT | | | One-Shot CoT | | | ILP Python | | | Greedy |
|---|---|---|---|---|---|---|---|---|---|---|---|---|---|---|
| | | GPT | Llama | Qwen | GPT | Llama | Qwen | GPT | Llama | Qwen | GPT | Llama | Qwen | |
| GCP | Textbook | 42.0 | 9.3 | 17.3 | 60.7 | 38.7 | 34.7 | 60.0 | 52.0 | 54.7 | 56.0 | 14.0 | 12.0 | 98.0 |
| | Inverted | −39.3 | +4.7 | −16.0 | −59.4 | −38.7 | −21.4 | −59.3 | −52.0 | −44.0 | −41.3 | −7.3 | −2.0 | |
| | Costumed | −6.2 | −6.2 | −3.7 | −6.5 | −17.8 | +16.4 | −4.7 | −19.6 | −1.4 | −43.8 | +20.7 | +5.6 | |
| KSP | Textbook | 22.7 | 15.3 | 14.0 | 48.0 | 37.3 | 29.3 | 50.0 | 37.3 | 36.7 | 89.3 | 51.3 | 45.3 | 75.3 |
| | Inverted | +4.6 | −7.3 | −5.3 | +2.7 | −2.6 | −1.3 | −4.7 | −26.0 | −23.4 | −0.6 | +6.0 | −11.3 | |
| | Costumed | −2.0 | −1.5 | −5.8 | −1.8 | −4.9 | −4.6 | −2.2 | −4.4 | −2.0 | −7.5 | −0.9 | −6.9 | |
| TSP | Textbook | 34.7 | 28.7 | 27.3 | 31.3 | 25.3 | 27.3 | 37.3 | 25.3 | 27.3 | 86.0 | 15.3 | 13.3 | 30.7 |
| | Inverted | −20.7 | −24.0 | −23.3 | −14.0 | −11.3 | −18.6 | −9.3 | −15.3 | −22.0 | −10.7 | −10.6 | −6.0 | |
| | Costumed | −8.3 | −14.0 | −13.7 | −1.7 | −5.5 | −15.1 | −9.1 | −8.0 | −5.5 | −37.1 | −11.5 | +4.7 | |

Table 1: Optimization accuracy on EHOP-RANDOM, broken down by problem variant. Values from the non-textbook variants are provided as their differences relative to Textbook. "Costumed" is the average over the three costumes of each base problem. The Greedy column shows the optimization accuracy for the greedy baselines.

ILP Python degrades the slowest with instance size. In this condition, the LLM is still required to make use of its "world knowledge" to flesh out the textual problem into a fine-grained symbolic ILP specification. However, it is freed up from having to perform complex combinatorial reasoning and keeping track of long chains of intermediate results (Zhang et al., 2024), which becomes exponentially harder as instances scale up. Unlike the other strategies, the ILP approach does not expose the LLM to the NP-hardness of the problem; the complexity of the language-to-ILP translation task grows linearly with input length.

### 5.2 Textbook is easier than other variants

As Table 1 shows, the methods we evaluated perform better on the Textbook variant than on the other variants in almost all conditions. The rows labeled "Inverted" represent the inverted Textbook variants; the "Costumed" rows are averages over all three costumes. Results for individual variants, including ones that are inverted *and* costumed at the same time, are in Appendix E. The drop is especially pronounced for the inverted problems, which are worded in ways that make them recognizably related to well-documented archetypes of NP-hard problems (see Section 5.5 for analysis).

While the ILP Python prompting strategy outperforms the others, it is still sensitive to deviations from the textbook presentations. This suggests that while the model no longer struggles to perform the right computation, the task of translating a problem to code is nevertheless affected by the ability to recognize the problem (when it is costumed) or to recognize how it deviates from the standard assumptions (when it is inverted).

The results for OPRO are not directly comparable with the numbers in Table 1 because we only ran OPRO on two instance sizes per variant. We find that OPRO achieves a higher overall optimization accuracy than CoT (e.g., 100% correct on textbook Traveling Salesman with 5 cities, 36% correct for 9 cities); details are in Table 13 in Appendix G. Nonetheless, Textbook is still easier than the other variants, with Inverted dropping by about 20 percentage points on average across all base problems and Costumed dropping by 8 points; the gap is wider for large instances. Thus, even a popular prompting strategy specifically developed for optimization is vulnerable to the presentation of the problem.

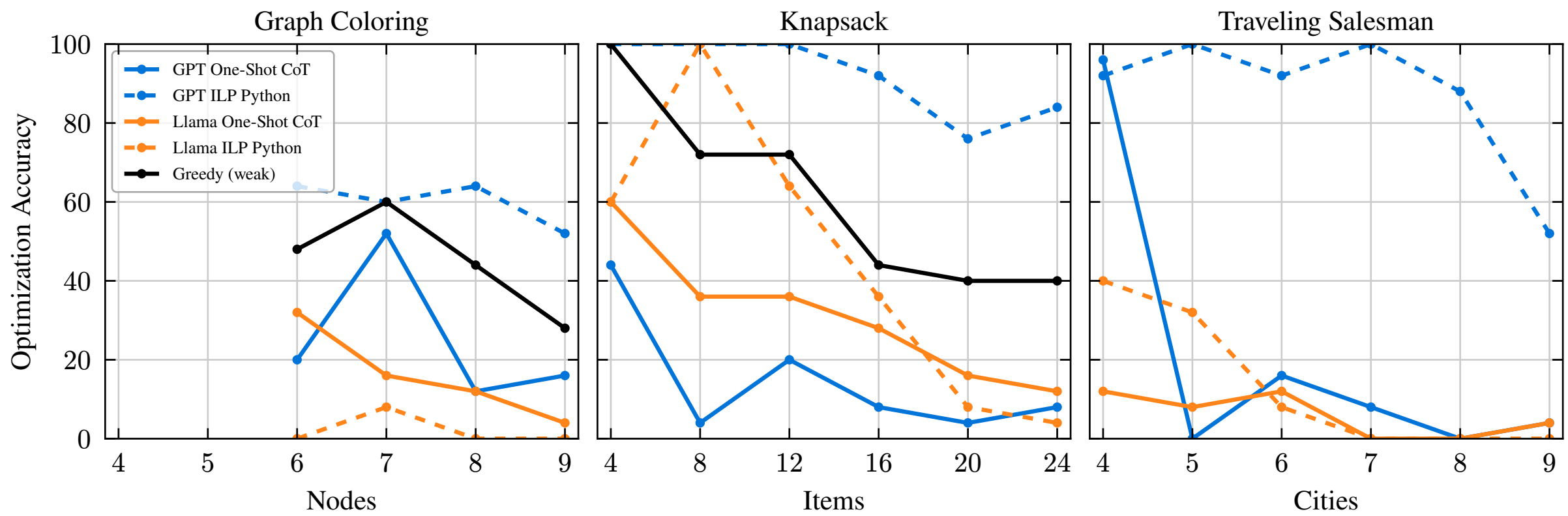

Figure 4: Optimization accuracy on the textbook variants in EHOP-HARD as a function of instance size. Note that this plot uses different greedy heuristics than Figure 3. See Appendix F for Qwen results.

| Problem | Variant | One-Shot | | | Zero-Shot CoT | | | One-Shot CoT | | | ILP Python | | | Greedy |
|---|---|---|---|---|---|---|---|---|---|---|---|---|---|---|
| | | GPT | Llama | Qwen | GPT | Llama | Qwen | GPT | Llama | Qwen | GPT | Llama | Qwen | |
| GCP | Textbook | 16.0 | 1.0 | 7.0 | 25.0 | 7.0 | 6.0 | 25.0 | 16.0 | 30.0 | 60.0 | 2.0 | 9.0 | 45.0 |
| | Inverted | −16.0 | +4.0 | −7.0 | −25.0 | −7.0 | −3.0 | −24.0 | −16.0 | −28.0 | −54.0 | −1.0 | +3.0 | |
| | Costumed | +5.3 | −1.0 | +1.0 | +0.7 | −1.0 | +21.3 | −0.7 | −7.0 | −6.3 | −52.7 | +19.3 | +1.7 | |
| KSP | Textbook | 8.7 | 5.3 | 9.3 | 18.0 | 10.7 | 26.0 | 14.7 | 31.3 | 11.3 | 92.0 | 45.3 | 37.3 | 61.3 |
| | Inverted | +11.3 | +9.4 | −2.6 | +18.7 | +15.3 | −5.3 | +24.6 | −17.3 | +0.0 | −4.7 | +8.0 | +1.4 | |
| | Costumed | +2.2 | +5.1 | +0.9 | +3.6 | +5.1 | −6.4 | +9.5 | −5.7 | +8.3 | −8.0 | −1.1 | −1.1 | |
| TSP | Textbook | 15.3 | 8.0 | 11.3 | 24.7 | 12.0 | 13.3 | 20.7 | 6.0 | 17.3 | 87.3 | 13.3 | 7.3 | — |
| | Inverted | −4.6 | −6.7 | −3.3 | −6.7 | −5.3 | −3.3 | −4.7 | −2.7 | −11.3 | −12.6 | −7.3 | +0.0 | |
| | Costumed | −1.7 | −3.3 | −2.9 | −3.6 | −4.4 | −4.4 | −11.6 | −0.7 | −6.6 | −33.7 | −9.5 | +10.3 | |

Table 2: Optimization accuracy on EHOP-HARD, broken down by problem variant. Formatted as in Table 1.

### 5.3 LLMs rarely beat greedy heuristics

One of the most striking findings of Figure 3 is the extent to which the greedy heuristics are competitive with the LLM-based approaches: the greedy approach is near-optimal on GRAPH COLORING, outperforms CoT reasoning on KNAPSACK, and is on par with it on TRAVELING SALESMAN. This raises the question of whether the LLM-based solvers achieve their relatively high accuracies in Table 1 only because the instances in EHOP-RANDOM are easy for their size.

We analyze the exact impact of instance difficulty on the performance of the different strategies by constructing a second sub-dataset of EHOP, which we call EHOP-HARD. This dataset is generated similarly to EHOP-RANDOM, except we only use instances which the greedy heuristics of Section 4.2 do not solve optimally. This results in the GRAPH COLORING instances being limited to instance sizes 6–9, as virtually all instances with four or five nodes are solved optimally by the greedy heuristic (cf. Table 1).

We repeat the analyses of Section 5.1 and Section 5.2 on EHOP-HARD. The results are shown in Figure 4 and Table 2. Note that we use a different set of **weak greedy heuristics** than in Figure 3, because EHOP-HARD is constructed such that the original greedy heuristics solve none of the instances optimally. Specifically, for GRAPH COLORING, we color the nodes in random order, rather than in descending order of degree; for KNAPSACK, we pick the highest-value, rather than the highest-density, items first. We call these heuristics "weak" because they performed worse than the original heuristics on EHOP-RANDOM.

The purely LLM-based approaches perform much worse overall than in the experiments on EHOP-RANDOM, giving further evidence to the interpretation that they primarily follow a greedy strategy. While their accuracy does not drop to zero, they are still being systematically outperformed by greedy heuristics. ILP Python performs similarly on HARD and RANDOM, illustrating the strength of the translation-based method. The overall pattern in Table 2 is still that the Textbook variant is easier than the others, except for methods that already perform very poorly on Textbook (see Appendix K for further discussion of deviations from this trend).

| Problem | Variant | EHOP-RANDOM | | | | EHOP-HARD | | | |
|---|---|---|---|---|---|---|---|---|---|
| | | Zero-Shot | | ILP Python | | Zero-Shot | | ILP Python | |
| | | R1 | Qwen | R1 | Qwen | R1 | Qwen | R1 | Qwen |
| GCP | Textbook | 100.0 | 77.3 | 91.3 | 76.0 | 98.0 | 62.0 | 94.0 | 72.0 |
| | Inverted | −62.0 | −42.6 | −40.0 | −0.7 | −75.0 | −50.0 | −56.0 | +1.0 |
| | Costumed | −2.9 | +0.3 | +5.8 | +2.9 | −4.0 | +11.3 | +3.3 | +6.3 |
| KSP | Textbook | 62.7 | 28.7 | 98.0 | 86.7 | 48.7 | 23.3 | 97.3 | 90.0 |
| | Inverted | +4.6 | −0.7 | +0.7 | −5.4 | +14.0 | −2.0 | +0.7 | −7.3 |
| | Costumed | −0.3 | −1.4 | +1.3 | −8.0 | +5.1 | −1.3 | +1.6 | −6.9 |
| TSP | Textbook | 34.7 | 14.7 | 82.0 | 56.0 | 32.0 | 14.0 | 72.7 | 50.7 |
| | Inverted | −6.7 | +2.0 | −10.7 | +0.0 | −0.7 | +3.3 | +8.6 | +4.6 |
| | Costumed | −12.0 | −0.9 | −5.6 | −12.2 | −10.7 | −2.2 | +4.2 | −4.7 |

Table 3: Optimization accuracy of DeepSeek-R1 and thinking-mode Qwen 3 (32B). Formatted as in Table 1.

### 5.4 Reasoning models are still sensitive to variation in presentation

The results for reasoning models are shown in Table 3, with more details in Table 14. We report the results in a separate table to emphasize that "zero-shot" prompting means something very different for reasoning models, given that they generate chains of thought even without being prompted to do so. While the trend across variants is not as clear for reasoning models, there is still a great deal of volatility, indicating that these models are similarly sensitive to the presentation of a problem.

An inspection of the reasoning traces reveals that DeepSeek-R1 frequently identifies the base problem: in both zero-shot and ILP conditions, R1 mentions a form of the problem name about 70% of the time (cf. Table 15). Even for Inverted Textbook, this is true for 69% of instances; in this case, R1 often includes the thought that the problem is *not* the textbook problem. Qwen3 recognizes the problem in about 64% of ILP problems, but only about 22% of the time in the zero-shot condition. In general, GRAPH COLORING is recognized the most, with the other two problems being recognized about equally often.

### 5.5 What do failures look like?

One tempting explanation for the poor performance of LLMs on the Inverted variants in particular is that they might not pay sufficient attention to the few tokens that distinguish Inverted from Textbook and attempt to solve the Textbook problem instead. We quantified this effect by counting the proportion of suboptimal and erroneous solutions of the Inverted instances that would have been optimal for Textbook with the same parameters. For GPT-4o, on average across the three base problems, this proportion is only 10% for One-Shot CoT; it is 16% for ILP Python (see Appendix I for details). This suggests that the problem on Inverted goes deeper than misreading a single token.

Given R1′s success with recognizing the base problem, one could assume that the gap between Textbook and variants could also be closed for the other LLMs by mentioning the base problem. We investigated this by prepending each prompt with "I am trying to solve a problem that I think resembles the <*base-problem*> problem." Adding such hints improved optimization accuracy compared to the condition without hints in a handful of cases (e.g. for One-Shot CoT with hints on KSP, the variants are even solved more accurately than Textbook), but in most conditions, the gap between Textbook and the variants persisted. Detailed results are in Table 17 in Appendix J.

### 5.6 Qualitative Analysis

To further our understanding of DeepSeek-R1′s error patterns, we manually review reasoning traces from 200 non-optimal responses: 100 from the zero-shot condition, and 100 from ILP Python. In the zero-shot condition, the vast majority (93%) of the reviewed reasoning traces reach the token limit before settling on a solution, resulting in a null response. R1 is clearly capable of solving some instances through reasoning output alone, but it often does so by generating and considering as many potential solutions as possible. This approach does not scale to NP-hard problems. Furthermore, R1 has a propensity for questioning and triple-checking every intermediate conclusion at which it arrives, padding its output with unnecessary re-computations and creating lengthy traces even for simple instances.

The ILP strategy avoids this pitfall, but encounters a variety of other issues. 41% of the reviewed traces involve R1 ignoring the instruction to cease its explanations after producing an answer, often generating additional code output or analysis of its solution which break the formatting requirements. 32% of the reviewed failures were due to Python errors, most often due to unclosed parentheses. Another 20% involved a failure in copying the specifications of the problem, often in the context of graph coloring instances where the model hallucinated or forgot edges in the graph.

## 6 Discussion

The results above paint an intricate picture of the features that make it easy or difficult for an LLM to solve NP-hard optimization problems. First, given previous research, it was expected that *instance size* would negatively impact accuracy. Second, we have identified *instance difficulty* as an important factor: among instances of the same size, those that cannot be solved by greedy heuristics are also harder for LLMs. Neurosymbolic methods that combine LLMs as semantic parsers with exact ILP solvers are more robust to both of these factors.

As our main result, we established that the *presentation* of a problem instance impacts how difficult it is for LLM-based methods to solve. Our "costumes" are quite transparent compared to the way we might expect a problem to be described by a lay user in real life. Even so, all standard LLMs perform much worse on our costumed and inverted variants compared to the well-established textbook presentations. This is true even when using advanced methods like OPRO. The story is a bit more nuanced for reasoning models, in which Textbook sometimes outperforms the variants and is sometimes outperformed by them. However, given that Textbook and the variants describe *the exact same underlying optimization problem* and the fact that the discrepancies persist even when the base problem is mentioned or recognized, our results are still evidence against a robust reasoning mechanism.

Furthermore, it is worth noting that DeepSeek-R1 solves the everyday variants by recognizing the underlying textbook problem and then using strategies for solving this textbook problem. This is a valid strategy for the costumes in EHOP, but many optimization problems that arise in real life—be it airline scheduling, allocating organ donations, or travel planning—are not just dressed-up versions of a single textbook problem. Thus, strong performance on EHOP does not ensure strong performance on real everyday problems.

## 7 Conclusion

We have shown that the ability of an LLM to solve NP-hard optimization problems is strongly affected by the presentation of the problem; there are reliable and large differences between the well-documented textbook form and the everyday problems we developed for this paper. At least standard LLMs, such as GPT-4o, seem to often recite when they appear to be reasoning.

One limitation of EHOP as a dataset of real problem-solving tasks is that real users will often not be able to spell out an instance of an everyday problem in detail, e.g. by assigning a numeric satisfaction value to every museum in Paris. It would be interesting to explore dialogue systems performing actual collaborative problem-solving with the user. The costumes of EHOP could be a good starting point for such work.

**Acknowledgments.** We gratefully acknowledge fruitful conversations with Peter Clark and the members of the Computational Linguistics group at Saarland University. This work was funded in part by the Deutsche Forschungsgemeinschaft (DFG, German Research Foundation) – GRK 2853/1 "Neuroexplicit Models of Language, Vision, and Action" - project number 471607914.

## Limitations

The instances of EHOP cover a limited range of instance sizes. The sizes for each base problem were informed by the performance of the greedy heuristics, and as we mentioned above, LLM optimization accuracy drops rapidly for larger instance sizes. As LLMs improve, it may become informative to evaluate on larger instances. We have made the code for generating more EHOP-like task descriptions available alongside the dataset itself to facilitate this.

We only included four LLMs in the evaluation (GPT-4o, Llama-3.1-70B Instruct, DeepSeek-R1, and Qwen3-32B), and we used a limited set of prompting strategies. Any research that uses finite compute resources will have this limitation. Nonetheless, we find very similar patterns on four very different strong models of different sizes, and on prompting strategies that span the range from very simple (one-shot without CoT) to the very complex (OPRO, R1′s reasoning-optimized thinking process). We thus believe that we can reasonably conclude that the generalization gap between Textbook and the other presentations is a real phenomenon that warrants further study.

Furthermore, EHOP is based on three well-established textbook problems, and the costumes do not actually cover full-blown real-world use cases like the ones in NL4Opt (Ramamonjison et al., 2022). This is because we did not construct EHOP to be predictive of real-world problem-solving accuracies but instead to permit a targeted comparison of the impact of problem presentation.

Finally, we have not compared the optimization accuracy of LLMs against that of humans. We have not included a human study because our focus was on the gap between Textbook and the other variants, not on the overall accuracy of the LLMs. For future work that aims to contextualize the general ability of an LLM to solve NP-hard optimization problems, a study with humans could be relevant.

## A Language Model Details

For both GPT-4o (`gpt-4o-2024-08-06`) and Llama-3.1-70B Instruct, we use the following sampling parameters for all LLM-only prompting strategies:

```
max_tokens=1024
temperature=0.0
presence_penalty=0.0
frequency_penalty=0.0
seed=1
```

In the case of the ILP LP prompting strategy, `max_tokens` is set to 6000 for the completion that is meant to produce the LP code. We similarly change `max_tokens` to 3072 for the ILP Python prompting strategy in the generation step. After the generation step, `max_tokens` is reset to 1024 (when asking the LLM to translate code output back to NL).

DeepSeek-R1 was tested using mostly the same parameters as the other models, with two main differences. We use the recommended temperature of 0.6 and scale up the token generation limit by a factor of 10 to allow reasoning traces to complete (the solution is always extracted from the final line of LLM output containing a list of numbers, so performance was near-zero without this extension).

Qwen3-32B was evaluated using the recommended sampling parameters for the thinking and non-thinking modes. This meant using `min_p=0` and `top_k=20`, as well as using a temperature of 0.6 and a `top_p` of 0.95 in thinking mode and 0.7 and 0.8 when in non-thinking mode. When thinking was enabled, we scaled up token limits by a factor of 4.

It is not known how many parameters GPT-4o has; Llama 3.1 70B Instruct has 70 billion parameters, DeepSeek-R1 has 671 billion, and Qwen3-32B has 32 billion. GPT-4o was prompted using API calls, so we do not know the GPU cost associated with running this subset of the experiments, though the API calls took about 50 hours in total to complete (excluding the ILP LP prompting strategy). We estimate that it takes about 240 GPU hours running on NVIDIA H100 PCIe GPUs to run the entire experiment (excluding ILP LP) on Llama-3.1-70B Instruct.

## B Result Category Examples

Table 5 shows examples of each result type (optimal, suboptimal, erroneous, incompatible, failure), within the context of TRAVELING SALESMAN instances with four or five cities. In each case, the optimal solution is shown in the rightmost column.

It should be noted that since models often would repeat the first node at the end of a tour (as seen in all of the responses in this table), we treated both "1, 2, 3, 4" and "1, 2, 3, 4, 1" as proper encodings of the round trip 1, 2, 3, 4, 1.

The response "1, 1, 1, 1, 1" is classified as *erroneous* since it has the right length (5 locations) but it does not meet the constraint of visiting each location exactly once. The response "1, 4, 1, 2, 3, 5, 1", on the other hand, is classified as *incompatible* since it has 7 locations (6 after removing the redundant 1 at the end), but the graph only has five nodes.

| Problem | Costume | Variant | Word Count |
|---|---|---|---|
| GCP | Textbook | Standard | $114 + 7n$ |
| | | Inverted | $114 + 7n$ |
| | Parties With Exes | Standard | $189 + 11n$ |
| | | Inverted | $193 + 11n$ |
| | Taekwondo Tournament | Standard | $174 + 10n$ |
| | | Inverted | $168 + 10n$ |
| | Student Groups | Standard | $146 + 7n$ |
| | | Inverted | $158 + 7n$ |
| KSP | Textbook | Standard | $82 + 14n$ |
| | | Inverted | $78 + 14n$ |
| | Lemonade Stand | Standard | $115 + 10n$ |
| | | Inverted | $118 + 10n$ |
| | Party Planning | Standard | $94 + 13n$ |
| | | Inverted | $121 + 13n$ |
| | Sightseeing | Standard | $119 + 13n$ |
| | | Inverted | $152 + 13n$ |
| TSP | Textbook | Standard | $70 + 9n$ |
| | | Inverted | $70 + 9n$ |
| | Exercise Schedule | Standard | $191 + 10n$ |
| | | Inverted | $191 + 10n$ |
| | Task Schedule | Standard | $170 + 12n$ |
| | | Inverted | $185 + 12n$ |
| | UN Seating | Standard | $173 + 9n$ |
| | | Inverted | $171 + 9n$ |

Table 4: Word Count of every prompt as a function of instance scale.

<table>
<tr><th>Result</th><th>LLM Response</th><th>Optimal Solution</th></tr>
<tr><td>Optimal</td><td>1,2,3,4,1</td><td>1, 2, 3, 4</td></tr>
<tr><td>Suboptimal</td><td>1,4,3,2,1</td><td>1, 4, 2, 3</td></tr>
<tr><td>Erroneous</td><td>1, 1, 1, 1, 1</td><td>1, 4, 3, 5, 2</td></tr>
<tr><td>Incompatible</td><td>1,4,1,2,3,5,1</td><td>1, 4, 2, 3, 5</td></tr>
<tr><td>ILP Failure</td><td>

```
from gurobipy import GRB, Model, quicksum

def f():
    # Create the model
    model = Model("Traveling Salesman Problem")

    # Create helper variables
    n = 4  # number of cities
    dist = [[0, 5, 11, 4], [5, 0, 1, 1], [11, 1, 0, 3], [4, 1, 3, 0]]  # distance matrix

    # Add variables
    x = model.addVars(n, n, vtype=GRB.BINARY, name="x")  # x[i, j] = 1 if we travel from city i to city j
    u = model.addVars(n, vtype=GRB.INTEGER, name="u")  # u[i] = order in which we visit city i

    # Add constraints
    model.addConstrs(quicksum(x[i, j] for j in range(n)) == 1 for i in range(n))  # each city is visited exactly once
    model.addConstrs(quicksum(x[j, i] for j in range(n)) == 1 for i in range(n))  # each city is left exactly once
    model.addConstrs(u[i] - u[j] + n * x[i, j] <= n - 1 for i in range(n) for j in range(n) if i != j)  # subtour elimination
    model.addConstrs(x[i, i] == 0 for i in range(n))  # we cannot visit the same city twice
    model.addConstr(u[0] == 1)  # we start at city 1

    # Set objective
    model.setObjective(quicksum(dist[i][j] * x[i, j] for i in range(n) for j in range(n)), GRB.MINIMIZE)

    # Optimize/solve the model
    model.optimize()

    # Return the optimized model
    return model
```

This ILP formulation uses the following variables:<br>...<br>The objective function is the total distance traveled, which is minimized.<br>

```
AttributeError at line 117: Unable to retrieve attribute 'X'
```

</td><td>1, 2, 3, 4</td></tr>
</table>

Table 5: The following examples are all generated by Llama for textbook TRAVELING SALESMAN with the ILP Python prompting strategy. Except for the code failure example, there was a code response which was then executed successfully and returned to the model before the final output was produced. The code which produced an error is shown in the ILP Failure case. The error here is indicative of an ILP model which cannot be properly optimized.

## C Costumes

Table 6, Table 7, and Table 8 display examples of how problem instances were presented to the LLM. The instances used to generate all examples were of the smallest size used in the EHOP dataset (4 nodes/4 items/4 cities).

We break down the size of these problem descriptions in Table 4. The table shows the size of the natural-language problem description in words, using the form $B + Sn$, where $n$ is the instance size and $B$ and $S$ are constants. For KNAPSACK, $n$ is the number of items in the knapsack. For TRAVELING SALESMAN and GRAPH COLORING, $n$ is the number of edges in the graph (unlike e.g. in Figure 3, where instance sizes are counted in nodes). Observe that the different variants of the same base problem do not differ a lot in length, especially given that the length of larger instances is dominated by the $Sn$ factor. This means that prompt length is not a factor that can explain the differences in optimization accuracy.

| | Standard | Inverted |
|---|---|---|
| Textbook | I have a network of 4 nodes, numbered 1 to 4, with various nodes being connected to one another. I want to color the nodes such that no two connected nodes have the same color.<br>The connections are as follows: Node 1 and node 3 are connected. Node 1 and node 4 are connected. Node 2 and node 3 are connected. Node 2 and node 4 are connected.<br>How can I color the nodes using the fewest colors possible? Generate a comma-separated list of the colors for each node, where the colors are represented by integers ranging from 1 to the number of colors used. The colors should be in the order of the vertices, so the first color will correspond to node 1, the second color will correspond to node 2, and so on. | I have a network of 4 nodes, numbered 1 to 4, with various nodes being connected to one another. I want to color the nodes such that no two unconnected nodes have the same color.<br>The connections are as follows: Node 1 and node 2 are connected. Node 3 and node 4 are connected.<br>How can I color the nodes using the fewest colors possible? Generate a comma-separated list of the colors for each node, where the colors are represented by integers ranging from 1 to the number of colors used. The colors should be in the order of the vertices, so the first color will correspond to node 1, the second color will correspond to node 2, and so on. |
| Student Groups | I am a teacher, and I want to assign my 4 students to different groups. I need the groups to focus, so I need to make sure that no two students who are friends with one another are in the same group, otherwise they may get distracted. I don't need the groups to all be the same size, but I want to minimize the total number of groups.<br>The friendships are as follows: Student 1 and student 3 are friends. Student 1 and student 4 are friends. Student 2 and student 3 are friends. Student 2 and student 4 are friends.<br>Which group should each student be assigned to? Generate a comma-separated list with each student's group, where the groups are represented by integers ranging from 1 to the total number of groups. The groups should be in the order of the students' numbers, so the first group in the list will correspond to student 1, the second group will correspond to student 2, and so on. | I am a teacher, and I want to assign my 4 students to different groups. I want the groups to have fun, so I need to make sure that only students who are friends with one another are in the same group. In other words, no group can have a pair of students who aren't friends with each other. I don't need the groups to all be the same size, but I want to minimize the total number of groups.<br>The friendships are as follows: Student 1 and student 2 are friends. Student 3 and student 4 are friends.<br>Which group should each student be assigned to? Generate a comma-separated list with each student's group, where the groups are represented by integers ranging from 1 to the total number of groups. The groups should be in the order of the students' numbers, so the first group in the list will correspond to student 1, the second group will correspond to student 2, and so on. |
| Parties with Exes | My birthday is coming up, and I want to celebrate with my 4 friends. Unfortunately, some of my friends used to be in romantic relationships with each other, and they don't get along anymore. I will therefore be having multiple birthday parties. I want to invite each person to one party, and I want to invite exes to different parties so that no two people who used to date one another are at the same party. I have a list of who used to date whom, and I want to host as few parties as possible while avoiding the awkwardness of having a pair of exes at the same party.<br>The past relationships are as follows: Friend 1 and friend 3 used to be in a relationship. Friend 1 and friend 4 used to be in a relationship. Friend 2 and friend 3 used to be in a relationship. Friend 2 and friend 4 used to be in a relationship.<br>Which party should each friend be invited to? Generate a comma-separated list with each friend's party, where the parties are represented by integers ranging from 1 to the total number of parties. The parties should be in the order of the friends' numbers, so the first party in the list will correspond to friend 1, the second party will correspond to friend 2, and so on. | My birthday is coming up, and I want to celebrate with my 4 friends. Some of my friends used to be in romantic relationships with each other, and they don't get along anymore. I will therefore be having multiple birthday parties. I want to invite each person to one party, and I want to make things as awkward as possible, so I only want to invite two people to the same party if they used to be in a relationship. I have a list of who used to date whom, and I want to host as few parties as possible while avoiding having a pair of people who haven't dated at the same party.<br>The past relationships are as follows: Friend 1 and friend 2 used to be in a relationship. Friend 3 and friend 4 used to be in a relationship.<br>Which party should each friend be invited to? Generate a comma-separated list with each friend's party, where the parties are represented by integers ranging from 1 to the total number of parties. The parties should be in the order of the friends' numbers, so the first party in the list will correspond to friend 1, the second party will correspond to friend 2, and so on. |
| Taekwondo Tournament | I am organizing a taekwondo tournament. There are 4 participants, and I need to reserve some rooms in the tournament hall for them to warm up in. I want to make sure that no two participants who are competing against each other are in the same room. This way, no one will learn about an opponent's technique ahead of the actual competition. I have a list of who is competing against whom, and I want to reserve as few rooms as possible while making sure no one is in the same room as any of their opponents.<br>Here are the matchups: Participant 1 and participant 3 are competing against one another. Participant 1 and participant 4 are competing against one another. Participant 2 and participant 3 are competing against one another. Participant 2 and participant 4 are competing against one another.<br>Which room should each participant be assigned to? Generate a comma-separated list with each participant's room, where the rooms are represented by integers ranging from 1 to the total number of rooms. The rooms should be in the order of the participants' numbers, so the first room in the list will correspond to participant 1, the second room will correspond to participant 2, and so on. | I am organizing a taekwondo tournament. There are 4 participants, and I need to reserve some rooms in the tournament hall for them to warm up in. I want to make sure that if two participants are not competing against each other, then they are in different rooms. This way, competitive tension will be as high as possible. I have a list of who is competing against whom, and I want to reserve as few rooms as possible while making sure no one is in the same room as a non-opponent.<br>Here are the matchups: Participant 1 and participant 2 are competing against one another. Participant 3 and participant 4 are competing against one another.<br>Which room should each participant be assigned to? Generate a comma-separated list with each participant's room, where the rooms are represented by integers ranging from 1 to the total number of rooms. The rooms should be in the order of the participants' numbers, so the first room in the list will correspond to participant 1, the second room will correspond to participant 2, and so on. |

Table 6: Examples of the four GRAPH COLORING costumes, both standard (textbook rules) and inverted, all generated using the same problem instance.

## D Prompting Strategies

Table 9 presents the overall structure of each prompting strategy. The BASE PROMPT would be of the form of one of the examples seen in Appendix C. It is also worth noting that the DEMO PROMPT and DEMO GREEDY CoT were always formatted to match the variant of the BASE PROMPT.

| | ➡️ Standard | 🔄 Inverted |
|---|---|---|
| 🎒 Textbook | I am trying to fill a bag with valuable items. Each item has a weight and a value.<br>Here are the items I have: Item 1 has a weight of 1 kg and a value of 2 €. Item 2 has a weight of 1 kg and a value of 2 €. Item 3 has a weight of 3 kg and a value of 3 €. Item 4 has a weight of 3 kg and a value of 4 €.<br>Which items should I pack to get the most value possible while also making sure the total weight of the items does not exceed the bag's capacity of 1 kg? Generate a comma-separated list of the items I should put in the bag, where each item is represented by its number. | I am trying to fill a bag with worthless items. Each item has a weight and a value.<br>Here are the items I have: Item 1 has a weight of 1 kg and a value of 2 €. Item 2 has a weight of 1 kg and a value of 2 €. Item 3 has a weight of 3 kg and a value of 3 €. Item 4 has a weight of 3 kg and a value of 4 €.<br>Which items should I pack to get the least value possible while also making sure the total weight of the items is at least 7 kg? Generate a comma-separated list of the items I should put in the bag, where each item is represented by its number. |
| 🍋 Lemonade Stand | I am running a lemonade stand where I don't set a single price but rather let the customers make custom offers. Each customer is offering a specific amount of money for a specific amount of lemonade. Each offer is rigid, so I can only fulfill it exactly as stated or not fulfill it at all.<br>I have the following offers: Customer 1 is offering $2 for 1 gallon of lemonade. Customer 2 is offering $2 for 1 gallon of lemonade. Customer 3 is offering $3 for 3 gallons of lemonade. Customer 4 is offering $4 for 3 gallons of lemonade.<br>Which customers' offers should I take up to make my revenue as large as possible given that I can't sell more than 1 total gallons of lemonade? Generate a comma-separated list of the customers whose offers I should take up, where each customer is represented by their number. | I am running a lemonade stand where I don't set a single price but rather let the customers make custom offers. Each customer is offering a specific amount of money for a specific amount of lemonade. Each offer is rigid, so I can only fulfill it exactly as stated or not fulfill it at all.<br>I have the following offers: Customer 1 is offering $2 for 1 gallon of lemonade. Customer 2 is offering $2 for 1 gallon of lemonade. Customer 3 is offering $3 for 3 gallons of lemonade. Customer 4 is offering $4 for 3 gallons of lemonade.<br>I don't want to seem greedy. Which customers' offers should I take up to make my total revenue as small as possible while selling at least 7 gallons of lemonade? Generate a comma-separated list of the customers whose offers I should take up, where each customer is represented by their number. |
| 🏛️ Sightseeing | I am going to be visiting Paris tomorrow, and I want to make the most of my time there. I have a list of attractions I want to visit, but I don't have enough time to visit all of them. I have given each attraction a point value and determined how many minutes I would need to spend on it.<br>Here are the attractions: Attraction 1 has a score of 2 points and would require 10 minutes. Attraction 2 has a score of 2 points and would require 10 minutes. Attraction 3 has a score of 3 points and would require 30 minutes. Attraction 4 has a score of 4 points and would require 30 minutes.<br>Which attractions should I visit to make the total point value as high as possible while not having the total time required go over my sightseeing limit of 10 minutes? Generate a comma-separated list of the attractions I should visit, where each attraction is represented by its number. | I am going to be visiting Paris tomorrow with a friend. I need to go through some emails at the start of the trip while my friend gets a head start on the sightseeing. I want to tell him which attractions he can visit before I join him so that I miss out as little as possible. I have given each attraction on our list a point value and determined how many minutes one would need to spend on it.<br>Here are the attractions: Attraction 1 has a score of 2 points and would require 10 minutes. Attraction 2 has a score of 2 points and would require 10 minutes. Attraction 3 has a score of 3 points and would require 30 minutes. Attraction 4 has a score of 4 points and would require 30 minutes.<br>Which attractions should I tell my friend to visit to make the total score of the attractions he sees without me as low as possible while ensuring that the total time required to visit them is at least 70 minutes? Generate a comma-separated list of the attractions I should suggest to my friend, where each attraction is represented by its number. |
| 🎉 Party Planning | I am planning a party, and I need to buy some decorations. Each decoration has a cost and a point value I've assigned in terms of its worth as a decoration.<br>Here are the decorations I can buy: Decoration 1 has a cost of $10 and a point value of 2. Decoration 2 has a cost of $10 and a point value of 2. Decoration 3 has a cost of $30 and a point value of 3. Decoration 4 has a cost of $30 and a point value of 4.<br>I can buy at most one of each decoration. Which decorations should I purchase to make the total point value as high as possible without going over my budget of $10? Generate a comma-separated list of the decorations I should buy, where each decoration is represented by its number. | I am planning a party, and I need to buy some decorations. I don't want the decorations to be the focus of the party, so I wan't to pick the worst ones, but I still need to spend the decorations budget. Each decoration has a cost and a point value I've assigned in terms of its worth as a decoration.<br>Here are the decorations I can buy: Decoration 1 has a cost of $10 and a point value of 2. Decoration 2 has a cost of $10 and a point value of 2. Decoration 3 has a cost of $30 and a point value of 3. Decoration 4 has a cost of $30 and a point value of 4.<br>I can buy at most one of each decoration. Which decorations should I purchase to make the total point value as low as possible while spending at least $70? Generate a comma-separated list of the decorations I should buy, where each decoration is represented by its number. |

Table 7: Examples of the four KNAPSACK costumes, both standard (textbook rules) and inverted, all generated using the same problem instance.

In the One-Shot strategies, we ensured that the example is from the same variant and of the largest input size for the base problem, e.g., a 9-node graph for all GRAPH COLORING instances. This ensures that any reduction in problem-solving accuracy is not caused by length generalization issues, which are a known problem for transformers (Zhou et al., 2024; Anil et al., 2022).

1. **Zero-Shot Chain-of-Thought (CoT)**: The task description is followed by the sentence "Let's think step by step." (Kojima et al., 2022)
2. **One-Shot Chain-of-Thought (CoT)**: We prepend to the prompt the same example used in the one-shot case, this time with an answer text that includes a chain of thought resulting in a solution (Wei et al., 2022).

In the one-shot strategies, the `Assistant` response was provided by us to emulate a past response in the conversational context. In the ILP cases, on the other hand, the `Assistant` response was in fact

| | Standard | Inverted |
|---|---|---|
| Textbook | I am planning a trip to visit several cities. Here are the distances between each pair of cities:<br>City 1 and city 2 are 8 miles apart. City 1 and city 3 are 14 miles apart. City 1 and city 4 are 13 miles apart. City 2 and city 3 are 6 miles apart. City 2 and city 4 are 15 miles apart. City 3 and city 4 are 3 miles apart.<br>What is the shortest possible route that starts at city 1, visits each city exactly once, and returns to city 1? Please generate a comma-separated list of the cities in the order I should visit them, where the cities are represented by their respective numbers. | I am planning a trip to visit several cities. Here are the distances between each pair of cities:<br>City 1 and city 2 are 11 miles apart. City 1 and city 3 are 5 miles apart. City 1 and city 4 are 6 miles apart. City 2 and city 3 are 13 miles apart. City 2 and city 4 are 4 miles apart. City 3 and city 4 are 16 miles apart.<br>What is the longest possible route that starts at city 1, visits each city exactly once, and returns to city 1? Please generate a comma-separated list of the cities in the order I should visit them, where the cities are represented by their respective numbers. |
| Task Schedule | I have a set of tasks that I have to complete every day. My boss always makes me start with task 1, but the order in which I complete the rest is up to me. It takes me a certain amount of time to modify my workspace to transition from one task to another, and at the end of the day, I'll need to set up my space for task 1 so that I'm ready the next morning. Here is the time it takes me to transition from one task to another:<br>It takes 8 minutes to transition between task 1 and task 2. It takes 14 minutes to transition between task 1 and task 3. It takes 13 minutes to transition between task 1 and task 4. It takes 6 minutes to transition between task 2 and task 3. It takes 15 minutes to transition between task 2 and task 4. It takes 3 minutes to transition between task 3 and task 4.<br>It takes me the same amount of time to transition between one task and another, regardless of which task I'm transitioning from and which task I'm transitioning to. In what order should I complete the tasks every day to minimize the total time spent transitioning between tasks? Please generate a comma-separated list of the tasks in the order I should complete them, where the tasks are represented by their respective numbers. | I have a set of tasks that I have to complete every day. My boss always makes me start with task 1, but the order in which I complete the rest is up to me. It takes me a certain amount of time to modify my workspace to transition from one task to another, and at the end of the day, I'll need to set up my space for task 1 so that I'm ready the next morning. Here is the time it takes me to transition from one task to another:<br>It takes 11 minutes to transition between task 1 and task 2. It takes 5 minutes to transition between task 1 and task 3. It takes 6 minutes to transition between task 1 and task 4. It takes 13 minutes to transition between task 2 and task 3. It takes 4 minutes to transition between task 2 and task 4. It takes 16 minutes to transition between task 3 and task 4.<br>It takes me the same amount of time to transition between one task and another, regardless of which task I'm transitioning from and which task I'm transitioning to, and the only time I get to relax during the day is during these transitions. In what order should I complete the tasks every day to maximize the total time spent transitioning between tasks? Please generate a comma-separated list of the tasks in the order I should complete them, where the tasks are represented by their respective numbers. |
| Exercise Schedule | My New Year's resolution is to be more physically active. I've made a list of 4 activities, and I want to do one of them every day. After I do an activity, I can't do it again until I've done everything else on the list. I'm going to start with activity 1 on January first, but the order in which I complete the rest is up in the air. Then, when I'm done with the list, I want to go through the activities again in the same order I used before. I've scored each pair of activities based on how similar they are, with more similar activities getting higher scores. Here are the scores:<br>Activity 1 and activity 2 have a similarity of 8. Activity 1 and activity 3 have a similarity of 14. Activity 1 and activity 4 have a similarity of 13. Activity 2 and activity 3 have a similarity of 6. Activity 2 and activity 4 have a similarity of 15. Activity 3 and activity 4 have a similarity of 3.<br>I want to have a lot of variety from day to day. What is the best order in which to do the activities to minimize the total similarity between activities on adjacent days, including between the last activity and activity 1 (when starting the next round)? Please generate a comma-separated list of the activities in the order I should complete them, where the activities are represented by their respective numbers. | My New Year's resolution is to be more physically active. I've made a list of 4 activities, and I want to do one of them every day. After I do an activity, I can't do it again until I've done everything else on the list. I'm going to start with activity 1 on January first, but the order in which I complete the rest is up in the air. Then, when I'm done with the list, I want to go through the activities again in the same order I used before. I've scored each pair of activities based on how similar they are, with more similar activities getting higher scores. Here are the scores:<br>Activity 1 and activity 2 have a similarity of 11. Activity 1 and activity 3 have a similarity of 5. Activity 1 and activity 4 have a similarity of 6. Activity 2 and activity 3 have a similarity of 13. Activity 2 and activity 4 have a similarity of 4. Activity 3 and activity 4 have a similarity of 16.<br>I want to have smooth transitions from one day to the next. What is the best order in which to do the activities to maximize the total similarity between activities on adjacent days, including between the last activity and activity 1 (when starting the next round)? Please generate a comma-separated list of the activities in the order I should complete them, where the activities are represented by their respective numbers. |
| UN Seating | I am responsible for the seating assignments at an upcoming UN meeting. There will be representatives from 4 nations sitting at a round table. The representative from nation 1 will be leading the discussion, so they will be sitting in the designated "Director Seat," but nothing else is decided yet. There is some amount of political tension between each pair of nations, and I've been given a list of tension scores for each pair of representatives, with higher scores indicating higher tension. Here are the tension levels between each pair of representatives:<br>Representative 1 and representative 2 have tension score 8. Representative 1 and representative 3 have tension score 14. Representative 1 and representative 4 have tension score 13. Representative 2 and representative 3 have tension score 6. Representative 2 and representative 4 have tension score 15. Representative 3 and representative 4 have tension score 3.<br>I want to minimize the total tension between adjacent pairs of representatives to prevent the discussion from getting heated. What should the seating order be, starting at the Director Seat and continuing clockwise? Note that the last person in the ordering will also be sitting next to the Director Seat. Please generate a comma-separated list of the representatives in the order they should be seated, where the representatives are represented by their respective numbers. | I am responsible for the seating assignments at an upcoming UN meeting. There will be representatives from 4 nations sitting at a round table. The representative from nation 1 will be leading the discussion, so they will be sitting in the designated "Director Seat," but nothing else is decided yet. There is some amount of political tension between each pair of nations, and I've been given a list of tension scores for each pair of representatives, with higher scores indicating higher tension. Here are the tension levels between each pair of representatives:<br>Representative 1 and representative 2 have tension score 11. Representative 1 and representative 3 have tension score 5. Representative 1 and representative 4 have tension score 6. Representative 2 and representative 3 have tension score 13. Representative 2 and representative 4 have tension score 4. Representative 3 and representative 4 have tension score 16.<br>I want to maximize the total tension between adjacent pairs of representatives to encourage discussion and progress. What should the seating order be, starting at the "Director Seat" and continuing clockwise? Note that the last person in the ordering will also be sitting next to the Director Seat. Please generate a comma-separated list of the representatives in the order they should be seated, where the representatives are represented by their respective numbers. |

Table 8: Examples of the four TRAVELING SALESMAN costumes, both standard (textbook rules) and inverted, all generated using the same problem instance.

| | | |
|---|---|---|
| Zero-Shot | `User:` | <BASE PROMPT><br>Please add no formatting and no explanations. |
| Zero-Shot CoT | `User:` | <BASE PROMPT><br>You may explain your reasoning, but do not add any more explanations once you have produced the comma-separated list.<br>Let's think step by step. |
| One-Shot | `User:` | <DEMO PROMPT> |
| | `Assitant:` | <DEMO ANSWER> |
| | `User:` | <BASE PROMPT> |
| One-Shot CoT | `User:` | <DEMO PROMPT> |
| | `Assitant:` | <DEMO GREEDY CoT><br><DEMO ANSWER> |
| | `User:` | <BASE PROMPT> |
| ILP LP | `User:` | <BASE PROMPT><br>Instead of solving the problem, please express it as an Integer Linear Programming (ILP) problem in the LP file format. Here is an example of the LP file format:<br>*LP EXAMPLE*<br>Start by thinking step by step about the variables and constraints you'll need in order to express the problem fully, and then create the specification in the LP format.<br><CAUTION AGAINST COMMON MISTAKES><br>Please provide the ILP problem in the LP format and do not solve the problem yourself. |
| | `Assistant:` | **<LLM GENERATED CODE>** |
| | `User:` | Your ILP problem was successfully solved. Here is the solution:<br><ILP MODEL PARAMETER VALUES><br>Translate this solution back to the original problem and provide it as originally specified.<br>Do not add any more explanation once you've provided the solution. |
| ILP Python | `User:` | <BASE PROMPT><br>Please express this as an Integer Linear Programming (ILP) problem using Python with the gurobipy library. Specifically, define a function named f that returns an optimized \`gurobipy.Model\` object which represents the problem. Here is an example of the format you should use for your answer:<br>*PYTHON EXAMPLE*<br>Start by thinking step by step about the variables and constraints you'll need in order to express the problem fully, and then define the Python function f.<br><CAUTION AGAINST COMMON MISTAKES> |
| | `Assistant:` | **<LLM GENERATED CODE>** |
| | `User:` | Your code was executed successfully. Here are all the variables of the model and their optimal values:<br><ILP MODEL PARAMETER VALUES><br>Translate this solution back to the original problem and provide it as originally specified.<br>Do not add any more explanation once you've provided the solution. |

Table 9: The structures of each prompting strategy.

| | | | | One-Shot | | | | Zero-Shot CoT | | | | One-Shot CoT | | | | ILP LP | | | | | ILP Python | | | | |
|---|---|---|---|---|---|---|---|---|---|---|---|---|---|---|---|---|---|---|---|---|---|---|---|---|---|
| | | | | O | S | E | I | O | S | E | I | O | S | E | I | O | S | E | I | F | O | S | E | I | F |
| RANDOM | 🖌️ GCP | ➡️ | 🖌️ | 42 | 9.3 | 48.7 | 0 | 60.7 | 4 | 34.7 | 0.7 | 60 | 2.7 | 37.3 | 0 | 42 | 7.3 | 48 | 0 | 2.7 | 56 | 14 | 25.3 | 4.7 | 0 |
| | | | 🏫 | 37.3 | 10.7 | 52 | 0 | 55.3 | 9.3 | 34.7 | 0.7 | 57.3 | 5.3 | 37.3 | 0 | 38 | 6.7 | 54.7 | 0.7 | 0 | 26 | 46 | 24 | 0.7 | 3.3 |
| | | | 💔 | 38.7 | 4.7 | 56.7 | 0 | 54 | 6 | 38 | 2 | 52 | 4 | 43.3 | 0.7 | 44.7 | 18.7 | 26.7 | 3.3 | 6.7 | 10 | 51.3 | 25.3 | 1.3 | 12 |
| | | | 🥋 | 31.3 | 18.7 | 50 | 0 | 53.3 | 14 | 30 | 2.7 | 56.7 | 3.3 | 40 | 0 | 19.3 | 13.3 | 58 | 0.7 | 8.7 | 0.7 | 0 | 0.7 | 0 | 98.7 |
| | | 🔄 | 🖌️ | 2.7 | 1.3 | 96 | 0 | 1.3 | 5.3 | 90.7 | 2.7 | 0.7 | 4.7 | 94.7 | 0 | 17.3 | 10 | 65.3 | 0 | 7.3 | 14.7 | 5.3 | 68.7 | 8 | 3.3 |
| | | | 🏫 | 27.3 | 8 | 64.7 | 0 | 46 | 4 | 50 | 0 | 47.3 | 8 | 44.7 | 0 | 10 | 4.7 | 80.7 | 0 | 4.7 | 40.7 | 19.3 | 32.7 | 5.3 | 2 |
| | | | 💔 | 22 | 9.3 | 68.7 | 0 | 15.3 | 8 | 74 | 2.7 | 26.7 | 10 | 63.3 | 0 | 18 | 15.3 | 50.7 | 4.7 | 11.3 | 34 | 29.3 | 27.3 | 4.7 | 4.7 |
| | | | 🥋 | 10 | 6.7 | 83.3 | 0 | 4 | 2 | 93.3 | 0.7 | 14 | 6.7 | 79.3 | 0 | 7.3 | 18.7 | 68 | 2.7 | 3.3 | 0 | 0 | 8.7 | 10 | 81.3 |
| | 🎒 KSP | ➡️ | 🎒 | 22.7 | 68 | 9.3 | 0 | 48 | 44 | 2 | 6 | 50 | 35.3 | 14 | 0.7 | 98.7 | 0.7 | 0.7 | 0 | 0 | 89.3 | 3.3 | 7.3 | 0 | 0 |
| | | | 🍋 | 23.3 | 63.3 | 13.3 | 0 | 49.3 | 35.3 | 13.3 | 2 | 52.7 | 35.3 | 10.7 | 1.3 | 99.3 | 0.7 | 0 | 0 | 0 | 84.7 | 5.3 | 10 | 0 | 0 |
| | | | 🏛️ | 21.3 | 72 | 6.7 | 0 | 45.3 | 49.3 | 5.3 | 0 | 48.7 | 42.7 | 7.3 | 1.3 | 99.3 | 0.7 | 0 | 0 | 0 | 76.7 | 7.3 | 16 | 0 | 0 |
| | | | 🎉 | 17.3 | 62.7 | 20 | 0 | 44 | 52 | 2.7 | 1.3 | 42 | 52 | 5.3 | 0.7 | 100 | 0 | 0 | 0 | 0 | 84 | 4 | 12 | 0 | 0 |
| | | 🔄 | 🎒 | 27.3 | 23.3 | 49.3 | 0 | 50.7 | 42.7 | 5.3 | 1.3 | 45.3 | 43.3 | 11.3 | 0 | 98 | 1.3 | 0.7 | 0 | 0 | 88.7 | 3.3 | 8 | 0 | 0 |
| | | | 🍋 | 12 | 37.3 | 50.7 | 0 | 52.7 | 38 | 8 | 1.3 | 47.3 | 40.7 | 12 | 0 | 99.3 | 0 | 0 | 0 | 0.7 | 78.7 | 7.3 | 14 | 0 | 0 |
| | | | 🏛️ | 9.3 | 23.3 | 67.3 | 0 | 27.3 | 50.7 | 17.3 | 4.7 | 24.7 | 58 | 17.3 | 0 | 98 | 0.7 | 1.3 | 0 | 0 | 74 | 8 | 18 | 0 | 0 |
| | | | 🎉 | 10 | 42 | 48 | 0 | 38 | 44.7 | 16.7 | 0.7 | 29.3 | 62 | 7.3 | 1.3 | 98 | 1.3 | 0.7 | 0 | 0 | 86 | 3.3 | 10.7 | 0 | 0 |
| | ✈️ TSP | ➡️ | ✈️ | 34.7 | 65.3 | 0 | 0 | 31.3 | 67.3 | 0.7 | 0.7 | 37.3 | 62.7 | 0 | 0 | 16.7 | 11.3 | 1.3 | 10 | 60.7 | 86 | 9.3 | 0 | 2.7 | 2 |
| | | | 📋 | 27.3 | 72.7 | 0 | 0 | 30.7 | 68.7 | 0 | 0.7 | 22.7 | 77.3 | 0 | 0 | 4.7 | 32.7 | 2.7 | 14 | 46 | 60 | 10 | 1.3 | 17.3 | 11.3 |
| | | | 📅 | 22.7 | 77.3 | 0 | 0 | 32 | 68 | 0 | 0 | 30.7 | 69.3 | 0 | 0 | 2 | 31.3 | 9.3 | 4 | 53.3 | 32.7 | 30.7 | 19.3 | 6 | 11.3 |
| | | | 🪑 | 29.3 | 70.7 | 0 | 0 | 26 | 72.7 | 0 | 1.3 | 31.3 | 68.7 | 0 | 0 | 2.7 | 40.7 | 0 | 0 | 56.7 | 54 | 44 | 0.7 | 0 | 1.3 |
| | | 🔄 | ✈️ | 14 | 86 | 0 | 0 | 17.3 | 68 | 2 | 12.7 | 28 | 72 | 0 | 0 | 14.7 | 6.7 | 0.7 | 6 | 72 | 75.3 | 20.7 | 1.3 | 2 | 0.7 |
| | | | 📋 | 14.7 | 85.3 | 0 | 0 | 24.7 | 75.3 | 0 | 0 | 18.7 | 80.7 | 0 | 0.7 | 2.7 | 14.7 | 1.3 | 6 | 75.3 | 46.7 | 22 | 2.7 | 18 | 10.7 |
| | | | 📅 | 30 | 70 | 0 | 0 | 34 | 63.3 | 0 | 2.7 | 28 | 72 | 0 | 0 | 2.7 | 11.3 | 12 | 4.7 | 69.3 | 16 | 12.7 | 44 | 11.3 | 16 |
| | | | 🪑 | 27.3 | 72.7 | 0 | 0 | 27.3 | 72.7 | 0 | 0 | 28 | 72 | 0 | 0 | 5.3 | 11.3 | 0 | 0.7 | 82.7 | 50.7 | 42 | 4.7 | 2 | 0.7 |
| HARD | 🖌️ GCP | ➡️ | 🖌️ | 16 | 15 | 69 | 0 | 25 | 18 | 53 | 4 | 25 | 14 | 61 | 0 | 40 | 5 | 49 | 0 | 6 | 60 | 7 | 30 | 3 | 0 |
| | | | 🏫 | 24 | 13 | 63 | 0 | 28 | 16 | 55 | 1 | 26 | 12 | 60 | 2 | 39 | 0 | 59 | 0 | 2 | 15 | 50 | 28 | 4 | 3 |
| | | | 💔 | 19 | 10 | 71 | 0 | 28 | 13 | 57 | 2 | 22 | 10 | 68 | 0 | 34 | 12 | 35 | 6 | 13 | 7 | 48 | 25 | 2 | 18 |
| | | | 🥋 | 21 | 22 | 57 | 0 | 21 | 31 | 46 | 2 | 25 | 9 | 66 | 0 | 20 | 6 | 68 | 3 | 3 | 0 | 0 | 1 | 0 | 99 |
| | | 🔄 | 🖌️ | 0 | 0 | 100 | 0 | 0 | 1 | 98 | 1 | 1 | 2 | 97 | 0 | 4 | 4 | 81 | 0 | 11 | 6 | 3 | 86 | 5 | 0 |
| | | | 🏫 | 8 | 8 | 84 | 0 | 23 | 9 | 68 | 0 | 33 | 15 | 52 | 0 | 4 | 1 | 94 | 0 | 1 | 42 | 10 | 36 | 11 | 1 |
| | | | 💔 | 6 | 7 | 87 | 0 | 2 | 5 | 93 | 0 | 3 | 18 | 77 | 2 | 10 | 11 | 58 | 10 | 11 | 37 | 34 | 28 | 1 | 0 |
| | | | 🥋 | 1 | 11 | 88 | 0 | 0 | 2 | 98 | 0 | 0 | 10 | 90 | 0 | 7 | 3 | 84 | 3 | 3 | 0 | 0 | 11 | 9 | 80 |
| | 🎒 KSP | ➡️ | 🎒 | 8.7 | 67.3 | 24 | 0 | 18 | 72 | 2 | 8 | 14.7 | 68.7 | 16 | 0.7 | 99.3 | 0 | 0.7 | 0 | 0 | 92 | 3.3 | 4.7 | 0 | 0 |
| | | | 🍋 | 11.3 | 66 | 22.7 | 0 | 14.7 | 60 | 21.3 | 4 | 24 | 60 | 14.7 | 1.3 | 100 | 0 | 0 | 0 | 0 | 82 | 6 | 12 | 0 | 0 |
| | | | 🏛️ | 8 | 77.3 | 14.7 | 0 | 22 | 74 | 2.7 | 1.3 | 16 | 72.7 | 10 | 1.3 | 98.7 | 0.7 | 0.7 | 0 | 0 | 84 | 6.7 | 9.3 | 0 | 0 |
| | | | 🎉 | 13.3 | 60 | 26.7 | 0 | 28 | 64 | 4 | 4 | 32.7 | 59.3 | 8 | 0 | 99.3 | 0.7 | 0 | 0 | 0 | 86 | 4 | 10 | 0 | 0 |
| | | 🔄 | 🎒 | 20 | 29.3 | 50.7 | 0 | 36.7 | 58.7 | 4 | 0.7 | 39.3 | 53.3 | 7.3 | 0 | 98.7 | 0.7 | 0.7 | 0 | 0 | 87.3 | 5.3 | 7.3 | 0 | 0 |
| | | | 🍋 | 14.7 | 37.3 | 48 | 0 | 30 | 60.7 | 8 | 1.3 | 26.7 | 64 | 8.7 | 0.7 | 98 | 0 | 1.3 | 0 | 0.7 | 77.3 | 6 | 14.7 | 2 | 0 |
| | | | 🏛️ | 13.3 | 14 | 72.7 | 0 | 26 | 54 | 15.3 | 4.7 | 28 | 62 | 10 | 0 | 98.7 | 0 | 1.3 | 0 | 0 | 78 | 5.3 | 16.7 | 0 | 0 |
| | | | 🎉 | 14 | 39.3 | 46.7 | 0 | 31.3 | 52.7 | 13.3 | 2.7 | 34.7 | 58 | 7.3 | 0 | 93.3 | 2.7 | 4 | 0 | 0 | 82 | 5.3 | 11.3 | 0.7 | 0.7 |
| | ✈️ TSP | ➡️ | ✈️ | 15.3 | 84.7 | 0 | 0 | 24.7 | 74 | 0 | 1.3 | 20.7 | 78 | 1.3 | 0 | 12.7 | 10.7 | 1.3 | 12.7 | 62.7 | 87.3 | 11.3 | 0.7 | 0.7 | 0 |
| | | | 📋 | 13.3 | 86.7 | 0 | 0 | 22.7 | 77.3 | 0 | 0 | 8 | 92 | 0 | 0 | 6 | 30 | 4 | 12 | 48 | 59.3 | 13.3 | 3.3 | 11.3 | 12.7 |
| | | | 📅 | 18 | 82 | 0 | 0 | 15.3 | 82.7 | 0.7 | 1.3 | 14 | 86 | 0 | 0 | 5.3 | 28.7 | 7.3 | 4.7 | 54 | 34.7 | 24.7 | 18.7 | 7.3 | 14.7 |
| | | | 🪑 | 9.3 | 90.7 | 0 | 0 | 25.3 | 74 | 0 | 0.7 | 5.3 | 94.7 | 0 | 0 | 3.3 | 40.7 | 0 | 0 | 56 | 66.7 | 29.3 | 4 | 0 | 0 |
| | | 🔄 | ✈️ | 10.7 | 89.3 | 0 | 0 | 18 | 70 | 0.7 | 11.3 | 16 | 84 | 0 | 0 | 14 | 6.7 | 0 | 6 | 73.3 | 74.7 | 20 | 3.3 | 2 | 0 |
| | | | 📋 | 8 | 92 | 0 | 0 | 21.3 | 76 | 0 | 2.7 | 8.7 | 90 | 0 | 1.3 | 1.3 | 7.3 | 3.3 | 3.3 | 84.7 | 35.3 | 26 | 2.7 | 21.3 | 14.7 |
| | | | 📅 | 8.7 | 91.3 | 0 | 0 | 15.3 | 82.7 | 0.7 | 1.3 | 3.3 | 96 | 0 | 0.7 | 3.3 | 8 | 15.3 | 2 | 71.3 | 19.3 | 14 | 40.7 | 12 | 14 |
| | | | 🪑 | 10.7 | 89.3 | 0 | 0 | 19.3 | 78 | 0 | 2.7 | 3.3 | 96.7 | 0 | 0 | 3.3 | 10.7 | 0.7 | 0 | 85.3 | 58.7 | 34.7 | 3.3 | 2 | 1.3 |

Table 10: Full results for GPT-4o on both EHOP-RANDOM and EHOP-HARD, including the ILP LP prompting strategy and a breakdown of result categories (➡️: standard, 🔄: inverted; O: optimal, S: suboptimal, E: erroneous, I: incompatible, F: ILP code failure). Costumes are represented by their emojis (established in Section 3). Greedy results do not vary by condition, and were provided in Table 1 and Table 2.

generated by the LLM, and the following `User` response would depend on its content. If the code

| | | | | One-Shot | | | | Zero-Shot CoT | | | | One-Shot CoT | | | | ILP LP | | | | | ILP Python | | | | |
|---|---|---|---|---|---|---|---|---|---|---|---|---|---|---|---|---|---|---|---|---|---|---|---|---|---|
| | | | | O | S | E | I | O | S | E | I | O | S | E | I | O | S | E | I | F | O | S | E | I | F |
| RANDOM | 🖌️ GCP | ➡️ | 🖌️ | 9.3 | 2.7 | 88 | 0 | 38.7 | 14 | 36.7 | 10.7 | 52 | 15.3 | 29.3 | 3.3 | 1.3 | 12.7 | 56 | 1.3 | 28.7 | 14 | 8.7 | 30.7 | 0 | 46.7 |
| | | | 🏫 | 0.7 | 4 | 95.3 | 0 | 21.3 | 42 | 30.7 | 6 | 28.7 | 35.3 | 32.7 | 3.3 | 1.3 | 11.3 | 48 | 0 | 39.3 | 38 | 6.7 | 44.7 | 2 | 8.7 |
| | | | 💔 | 4.7 | 0.7 | 94.7 | 0 | 18.7 | 9.3 | 49.3 | 22.7 | 34.7 | 16 | 42 | 7.3 | 4 | 8.7 | 25.3 | 32 | 30 | 26 | 10 | 45.3 | 7.3 | 11.3 |
| | | | 🥋 | 4 | 1.3 | 94 | 0.7 | 22.7 | 27.3 | 40 | 10 | 34 | 23.3 | 37.3 | 5.3 | 2 | 11.3 | 44.7 | 0 | 42 | 40 | 2.7 | 22.7 | 2 | 32.7 |
| | | 🔄 | 🖌️ | 14 | 2 | 84 | 0 | 0 | 2 | 90.7 | 7.3 | 0 | 3.3 | 86.7 | 10 | 1.3 | 8 | 50 | 0.7 | 40 | 6.7 | 3.3 | 59.3 | 0 | 30.7 |
| | | | 🏫 | 13.3 | 0 | 86.7 | 0 | 10 | 0 | 56.7 | 33.3 | 13.3 | 0 | 86 | 0.7 | 1.3 | 6 | 42 | 0 | 50.7 | 10 | 5.3 | 50 | 2 | 32.7 |
| | | | 💔 | 20 | 8.7 | 71.3 | 0 | 8 | 6 | 66 | 20 | 18 | 2 | 70.7 | 9.3 | 2 | 10 | 22 | 21.3 | 44.7 | 0 | 3.3 | 50 | 10.7 | 36 |
| | | | 🥋 | 19.3 | 3.3 | 77.3 | 0 | 8 | 4 | 78.7 | 9.3 | 11.3 | 2 | 79.3 | 7.3 | 0 | 6 | 57.3 | 0 | 36.7 | 0.7 | 0 | 26 | 0 | 73.3 |
| | 🎒 KSP | ➡️ | 🎒 | 15.3 | 58.7 | 24 | 2 | 37.3 | 42.7 | 6.7 | 13.3 | 37.3 | 50 | 12 | 0.7 | 92 | 6 | 2 | 0 | 0 | 51.3 | 18.7 | 29.3 | 0.7 | 0 |
| | | | 🍋 | 14 | 62.7 | 20 | 3.3 | 31.3 | 36.7 | 23.3 | 8.7 | 37.3 | 36.7 | 24 | 2 | 76.7 | 20.7 | 2.7 | 0 | 0 | 46 | 14.7 | 34 | 0 | 5.3 |
| | | | 🏛️ | 14.7 | 53.3 | 32 | 0 | 32.7 | 46.7 | 4.7 | 16 | 33.3 | 48.7 | 18 | 0 | 91.3 | 1.3 | 6 | 0 | 1.3 | 52 | 13.3 | 33.3 | 0 | 1.3 |
| | | | 🎉 | 12.7 | 48.7 | 38.7 | 0 | 33.3 | 50.7 | 8.7 | 7.3 | 28 | 47.3 | 24 | 0.7 | 94 | 5.3 | 0.7 | 0 | 0 | 53.3 | 11.3 | 35.3 | 0 | 0 |
| | | 🔄 | 🎒 | 8 | 24.7 | 56.7 | 10.7 | 34.7 | 39.3 | 22 | 4 | 11.3 | 46 | 34.7 | 8 | 90.7 | 0.7 | 5.3 | 0 | 3.3 | 57.3 | 4.7 | 38 | 0 | 0 |
| | | | 🍋 | 8.7 | 22 | 64 | 5.3 | 29.3 | 42 | 20.7 | 8 | 13.3 | 34.7 | 45.3 | 6.7 | 77.3 | 2.7 | 10.7 | 0 | 9.3 | 47.3 | 7.3 | 38.7 | 0.7 | 6 |
| | | | 🏛️ | 4.7 | 35.3 | 60 | 0 | 19.3 | 27.3 | 33.3 | 20 | 5.3 | 41.3 | 48 | 5.3 | 82.7 | 2.7 | 2 | 0 | 12.7 | 50 | 7.3 | 40 | 0 | 2.7 |
| | | | 🎉 | 2.7 | 42.7 | 50.7 | 4 | 20 | 39.3 | 24.7 | 16 | 9.3 | 36.7 | 45.3 | 8.7 | 90.7 | 1.3 | 7.3 | 0 | 0.7 | 33.3 | 6 | 60.7 | 0 | 0 |
| | ✈️ TSP | ➡️ | ✈️ | 28.7 | 71.3 | 0 | 0 | 25.3 | 52.7 | 1.3 | 20.7 | 25.3 | 74.7 | 0 | 0 | 0.7 | 2.7 | 0 | 1.3 | 95.3 | 15.3 | 33.3 | 14 | 6.7 | 30.7 |
| | | | 📋 | 18.7 | 81.3 | 0 | 0 | 23.3 | 61.3 | 0 | 15.3 | 19.3 | 80.7 | 0 | 0 | 0 | 1.3 | 0 | 0.7 | 98 | 7.3 | 16 | 4.7 | 7.3 | 64.7 |
| | | | 📅 | 8.7 | 91.3 | 0 | 0 | 17.3 | 74.7 | 2 | 6 | 12 | 87.3 | 0 | 0.7 | 0.7 | 5.3 | 2.7 | 3.3 | 88 | 4 | 18 | 0 | 0 | 78 |
| | | | 🪑 | 16.7 | 83.3 | 0 | 0 | 18.7 | 76.7 | 0 | 4.7 | 20.7 | 78 | 0 | 1.3 | 0.7 | 8.7 | 2.7 | 0 | 88 | 0 | 5.3 | 0 | 0 | 94.7 |
| | | 🔄 | ✈️ | 4.7 | 95.3 | 0 | 0 | 14 | 63.3 | 0 | 22.7 | 10 | 89.3 | 0 | 0.7 | 1.3 | 3.3 | 0 | 0.7 | 94.7 | 4.7 | 21.3 | 10 | 5.3 | 58.7 |
| | | | 📋 | 8 | 92 | 0 | 0 | 18 | 69.3 | 2 | 10.7 | 16.7 | 82.7 | 0 | 0.7 | 0 | 0.7 | 0.7 | 0 | 98.7 | 1.3 | 26.7 | 4 | 10 | 58 |
| | | | 📅 | 14 | 86 | 0 | 0 | 23.3 | 69.3 | 1.3 | 6 | 21.3 | 78.7 | 0 | 0 | 2 | 5.3 | 1.3 | 0.7 | 90.7 | 11.3 | 40.7 | 2.7 | 0 | 45.3 |
| | | | 🪑 | 14 | 86 | 0 | 0 | 20.7 | 76 | 0.7 | 2.7 | 22 | 78 | 0 | 0 | 0.7 | 0.7 | 3.3 | 0 | 95.3 | 0 | 0 | 0 | 0 | 100 |
| HARD | 🖌️ GCP | ➡️ | 🖌️ | 1 | 1 | 98 | 0 | 7 | 32 | 48 | 13 | 16 | 33 | 44 | 7 | 1 | 11 | 56 | 3 | 29 | 2 | 9 | 25 | 1 | 63 |
| | | | 🏫 | 0 | 0 | 100 | 0 | 9 | 40 | 40 | 11 | 5 | 55 | 39 | 1 | 0 | 12 | 46 | 0 | 42 | 31 | 4 | 59 | 3 | 3 |
| | | | 💔 | 0 | 2 | 97 | 1 | 4 | 9 | 57 | 30 | 13 | 34 | 47 | 6 | 0 | 4 | 23 | 40 | 33 | 9 | 11 | 56 | 8 | 16 |
| | | | 🥋 | 0 | 1 | 99 | 0 | 5 | 41 | 49 | 5 | 9 | 28 | 55 | 8 | 1 | 22 | 38 | 0 | 39 | 24 | 3 | 36 | 1 | 36 |
| | | 🔄 | 🖌️ | 5 | 2 | 93 | 0 | 0 | 0 | 83 | 17 | 0 | 4 | 83 | 13 | 0 | 3 | 43 | 4 | 50 | 1 | 5 | 47 | 1 | 46 |
| | | | 🏫 | 1 | 0 | 99 | 0 | 3 | 0 | 70 | 27 | 0 | 0 | 98 | 2 | 0 | 12 | 51 | 0 | 37 | 14 | 7 | 48 | 0 | 31 |
| | | | 💔 | 5 | 10 | 85 | 0 | 3 | 4 | 65 | 28 | 4 | 2 | 85 | 9 | 0 | 7 | 18 | 30 | 45 | 0 | 6 | 42 | 12 | 40 |
| | | | 🥋 | 5 | 2 | 93 | 0 | 2 | 7 | 80 | 11 | 0 | 2 | 95 | 3 | 0 | 4 | 59 | 1 | 36 | 0 | 0 | 11 | 2 | 87 |
| | 🎒 KSP | ➡️ | 🎒 | 5.3 | 68 | 25.3 | 1.3 | 10.7 | 72.7 | 4.7 | 12 | 31.3 | 60 | 7.3 | 1.3 | 92.7 | 6.7 | 0.7 | 0 | 0 | 45.3 | 19.3 | 35.3 | 0 | 0 |
| | | | 🍋 | 10.7 | 71.3 | 12 | 6 | 17.3 | 43.3 | 32 | 7.3 | 28.7 | 49.3 | 18.7 | 3.3 | 68.7 | 24 | 5.3 | 0.7 | 1.3 | 36.7 | 17.3 | 40.7 | 0 | 5.3 |
| | | | 🏛️ | 9.3 | 57.3 | 33.3 | 0 | 13.3 | 64 | 4.7 | 18 | 21.3 | 64 | 14.7 | 0 | 92 | 4 | 4 | 0 | 0 | 49.3 | 16 | 33.3 | 0 | 1.3 |
| | | | 🎉 | 11.3 | 52 | 36.7 | 0 | 16.7 | 68.7 | 6.7 | 8 | 26.7 | 50.7 | 22 | 0.7 | 92 | 6.7 | 1.3 | 0 | 0 | 46.7 | 10.7 | 42.7 | 0 | 0 |
| | | 🔄 | 🎒 | 14.7 | 18.7 | 54 | 12.7 | 26 | 54 | 14.7 | 5.3 | 14 | 42.7 | 36 | 7.3 | 88 | 0.7 | 7.3 | 0 | 4 | 53.3 | 11.3 | 35.3 | 0 | 0 |
| | | | 🍋 | 19.3 | 8.7 | 68.7 | 3.3 | 22 | 56.7 | 17.3 | 4 | 16.7 | 32 | 49.3 | 2 | 68.7 | 6.7 | 10.7 | 0 | 14 | 45.3 | 11.3 | 36.7 | 0.7 | 6 |
| | | | 🏛️ | 11.3 | 23.3 | 64.7 | 0.7 | 20.7 | 35.3 | 29.3 | 14.7 | 10 | 32.7 | 55.3 | 2 | 83.3 | 4.7 | 2.7 | 0 | 9.3 | 56 | 8 | 35.3 | 0 | 0.7 |
| | | | 🎉 | 8.7 | 28.7 | 54.7 | 8 | 26.7 | 32.7 | 26 | 14.7 | 13.3 | 33.3 | 45.3 | 8 | 87.3 | 0.7 | 10 | 0 | 2 | 36 | 14.7 | 49.3 | 0 | 0 |
| | ✈️ TSP | ➡️ | ✈️ | 8 | 92 | 0 | 0 | 12 | 62.7 | 1.3 | 24 | 6 | 94 | 0 | 0 | 1.3 | 2.7 | 0 | 0 | 96 | 13.3 | 36 | 8 | 6.7 | 36 |
| | | | 📋 | 5.3 | 94.7 | 0 | 0 | 7.3 | 72.7 | 0 | 20 | 6.7 | 93.3 | 0 | 0 | 0 | 2.7 | 0 | 0 | 97.3 | 8 | 18 | 2.7 | 9.3 | 62 |
| | | | 📅 | 3.3 | 96.7 | 0 | 0 | 9.3 | 80.7 | 2 | 8 | 4.7 | 94.7 | 0 | 0.7 | 0 | 4.7 | 2 | 2.7 | 90.7 | 3.3 | 20 | 0 | 0.7 | 76 |
| | | | 🪑 | 5.3 | 94.7 | 0 | 0 | 6 | 90.7 | 0 | 3.3 | 4.7 | 95.3 | 0 | 0 | 0.7 | 7.3 | 0.7 | 0 | 91.3 | 0 | 4.7 | 0 | 0 | 95.3 |
| | | 🔄 | ✈️ | 1.3 | 98.7 | 0 | 0 | 6.7 | 74 | 0.7 | 18.7 | 3.3 | 96 | 0 | 0.7 | 0.7 | 5.3 | 0 | 0.7 | 93.3 | 6 | 14 | 8 | 8.7 | 63.3 |
| | | | 📋 | 5.3 | 94.7 | 0 | 0 | 5.3 | 75.3 | 0.7 | 18.7 | 8 | 90.7 | 0 | 1.3 | 0 | 0 | 0.7 | 0.7 | 98.7 | 2.7 | 22 | 5.3 | 7.3 | 62.7 |
| | | | 📅 | 5.3 | 94.7 | 0 | 0 | 9.3 | 80.7 | 2.7 | 7.3 | 4.7 | 95.3 | 0 | 0 | 0 | 4 | 1.3 | 0 | 94.7 | 14 | 40 | 1.3 | 0.7 | 44 |
| | | | 🪑 | 7.3 | 92.7 | 0 | 0 | 6 | 90 | 0.7 | 3.3 | 4.7 | 95.3 | 0 | 0 | 0.7 | 2.7 | 1.3 | 0.7 | 94.7 | 0 | 0.7 | 0 | 0 | 99.3 |

Table 11: Full results for Llama-3.1-70B Instruct on both EHOP-RANDOM and EHOP-HARD, with formatting matching that of Table 10.

ran successfully, its output would be inserted in the format of the response shown, and if the code produced an error, the instance would be marked as a code failure, and there would be no follow-up. For full implementation details, see our codebase.

| | | | | One-Shot | | | | Zero-Shot CoT | | | | One-Shot CoT | | | | ILP LP | | | | | ILP Python | | | | |
|---|---|---|---|---|---|---|---|---|---|---|---|---|---|---|---|---|---|---|---|---|---|---|---|---|---|
| | | | | O | S | E | I | O | S | E | I | O | S | E | I | O | S | E | I | F | O | S | E | I | F |
| RANDOM | 🖌️ GCP | ➡️ | 🖌️ | 9.3 | 2.7 | 88 | 0 | 38.7 | 14 | 36.7 | 10.7 | 52 | 15.3 | 29.3 | 3.3 | 1.3 | 12.7 | 56 | 1.3 | 28.7 | 14 | 8.7 | 30.7 | 0 | 46.7 |
| | | | 🏫 | 0.7 | 4 | 95.3 | 0 | 21.3 | 42 | 30.7 | 6 | 28.7 | 35.3 | 32.7 | 3.3 | 1.3 | 11.3 | 48 | 0 | 39.3 | 38 | 6.7 | 44.7 | 2 | 8.7 |
| | | | 💔 | 4.7 | 0.7 | 94.7 | 0 | 18.7 | 9.3 | 49.3 | 22.7 | 34.7 | 16 | 42 | 7.3 | 4 | 8.7 | 25.3 | 32 | 30 | 26 | 10 | 45.3 | 7.3 | 11.3 |
| | | | 🥋 | 4 | 1.3 | 94 | 0.7 | 22.7 | 27.3 | 40 | 10 | 34 | 23.3 | 37.3 | 5.3 | 2 | 11.3 | 44.7 | 0 | 42 | 40 | 2.7 | 22.7 | 2 | 32.7 |
| | | 🔄 | 🖌️ | 14 | 2 | 84 | 0 | 0 | 2 | 90.7 | 7.3 | 0 | 3.3 | 86.7 | 10 | 1.3 | 8 | 50 | 0.7 | 40 | 6.7 | 3.3 | 59.3 | 0 | 30.7 |
| | | | 🏫 | 13.3 | 0 | 86.7 | 0 | 10 | 0 | 56.7 | 33.3 | 13.3 | 0 | 86 | 0.7 | 1.3 | 6 | 42 | 0 | 50.7 | 10 | 5.3 | 50 | 2 | 32.7 |
| | | | 💔 | 20 | 8.7 | 71.3 | 0 | 8 | 6 | 66 | 20 | 18 | 2 | 70.7 | 9.3 | 2 | 10 | 22 | 21.3 | 44.7 | 0 | 3.3 | 50 | 10.7 | 36 |
| | | | 🥋 | 19.3 | 3.3 | 77.3 | 0 | 8 | 4 | 78.7 | 9.3 | 11.3 | 2 | 79.3 | 7.3 | 0 | 6 | 57.3 | 0 | 36.7 | 0.7 | 0 | 26 | 0 | 73.3 |
| | 🫑 KSP | ➡️ | 🫑 | 15.3 | 58.7 | 24 | 2 | 37.3 | 42.7 | 6.7 | 13.3 | 37.3 | 50 | 12 | 0.7 | 92 | 6 | 2 | 0 | 0 | 51.3 | 18.7 | 29.3 | 0.7 | 0 |
| | | | 🍋 | 14 | 62.7 | 20 | 3.3 | 31.3 | 36.7 | 23.3 | 8.7 | 37.3 | 36.7 | 24 | 2 | 76.7 | 20.7 | 2.7 | 0 | 0 | 46 | 14.7 | 34 | 0 | 5.3 |
| | | | 🏛️ | 14.7 | 53.3 | 32 | 0 | 32.7 | 46.7 | 4.7 | 16 | 33.3 | 48.7 | 18 | 0 | 91.3 | 1.3 | 6 | 0 | 1.3 | 52 | 13.3 | 33.3 | 0 | 1.3 |
| | | | 🎉 | 12.7 | 48.7 | 38.7 | 0 | 33.3 | 50.7 | 8.7 | 7.3 | 28 | 47.3 | 24 | 0.7 | 94 | 5.3 | 0.7 | 0 | 0 | 53.3 | 11.3 | 35.3 | 0 | 0 |
| | | 🔄 | 🫑 | 8 | 24.7 | 56.7 | 10.7 | 34.7 | 39.3 | 22 | 4 | 11.3 | 46 | 34.7 | 8 | 90.7 | 0.7 | 5.3 | 0 | 3.3 | 57.3 | 4.7 | 38 | 0 | 0 |
| | | | 🍋 | 8.7 | 22 | 64 | 5.3 | 29.3 | 42 | 20.7 | 8 | 13.3 | 34.7 | 45.3 | 6.7 | 77.3 | 2.7 | 10.7 | 0 | 9.3 | 47.3 | 7.3 | 38.7 | 0.7 | 6 |
| | | | 🏛️ | 4.7 | 35.3 | 60 | 0 | 19.3 | 27.3 | 33.3 | 20 | 5.3 | 41.3 | 48 | 5.3 | 82.7 | 2.7 | 2 | 0 | 12.7 | 50 | 7.3 | 40 | 0 | 2.7 |
| | | | 🎉 | 2.7 | 42.7 | 50.7 | 4 | 20 | 39.3 | 24.7 | 16 | 9.3 | 36.7 | 45.3 | 8.7 | 90.7 | 1.3 | 7.3 | 0 | 0.7 | 33.3 | 6 | 60.7 | 0 | 0 |
| | ✈️ TSP | ➡️ | ✈️ | 28.7 | 71.3 | 0 | 0 | 25.3 | 52.7 | 1.3 | 20.7 | 25.3 | 74.7 | 0 | 0 | 0.7 | 2.7 | 0 | 1.3 | 95.3 | 15.3 | 33.3 | 14 | 6.7 | 30.7 |
| | | | 📋 | 18.7 | 81.3 | 0 | 0 | 23.3 | 61.3 | 0 | 15.3 | 19.3 | 80.7 | 0 | 0 | 0 | 1.3 | 0 | 0.7 | 98 | 7.3 | 16 | 4.7 | 7.3 | 64.7 |
| | | | 📅 | 8.7 | 91.3 | 0 | 0 | 17.3 | 74.7 | 2 | 6 | 12 | 87.3 | 0 | 0.7 | 0.7 | 5.3 | 2.7 | 3.3 | 88 | 4 | 18 | 0 | 0 | 78 |
| | | | 🪑 | 16.7 | 83.3 | 0 | 0 | 18.7 | 76.7 | 0 | 4.7 | 20.7 | 78 | 0 | 1.3 | 0.7 | 8.7 | 2.7 | 0 | 88 | 0 | 5.3 | 0 | 0 | 94.7 |
| | | 🔄 | ✈️ | 4.7 | 95.3 | 0 | 0 | 14 | 63.3 | 0 | 22.7 | 10 | 89.3 | 0 | 0.7 | 1.3 | 3.3 | 0 | 0.7 | 94.7 | 4.7 | 21.3 | 10 | 5.3 | 58.7 |
| | | | 📋 | 8 | 92 | 0 | 0 | 18 | 69.3 | 2 | 10.7 | 16.7 | 82.7 | 0 | 0.7 | 0 | 0.7 | 0.7 | 0 | 98.7 | 1.3 | 26.7 | 4 | 10 | 58 |
| | | | 📅 | 14 | 86 | 0 | 0 | 23.3 | 69.3 | 1.3 | 6 | 21.3 | 78.7 | 0 | 0 | 2 | 5.3 | 1.3 | 0.7 | 90.7 | 11.3 | 40.7 | 2.7 | 0 | 45.3 |
| | | | 🪑 | 14 | 86 | 0 | 0 | 20.7 | 76 | 0.7 | 2.7 | 22 | 78 | 0 | 0 | 0.7 | 0.7 | 3.3 | 0 | 95.3 | 0 | 0 | 0 | 0 | 100 |
| HARD | 🖌️ GCP | ➡️ | 🖌️ | 1 | 1 | 98 | 0 | 7 | 32 | 48 | 13 | 16 | 33 | 44 | 7 | 1 | 11 | 56 | 3 | 29 | 2 | 9 | 25 | 1 | 63 |
| | | | 🏫 | 0 | 0 | 100 | 0 | 9 | 40 | 40 | 11 | 5 | 55 | 39 | 1 | 0 | 12 | 46 | 0 | 42 | 31 | 4 | 59 | 3 | 3 |
| | | | 💔 | 0 | 2 | 97 | 1 | 4 | 9 | 57 | 30 | 13 | 34 | 47 | 6 | 0 | 4 | 23 | 40 | 33 | 9 | 11 | 56 | 8 | 16 |
| | | | 🥋 | 0 | 1 | 99 | 0 | 5 | 41 | 49 | 5 | 9 | 28 | 55 | 8 | 1 | 22 | 38 | 0 | 39 | 24 | 3 | 36 | 1 | 36 |
| | | 🔄 | 🖌️ | 5 | 2 | 93 | 0 | 0 | 0 | 83 | 17 | 0 | 4 | 83 | 13 | 0 | 3 | 43 | 4 | 50 | 1 | 5 | 47 | 1 | 46 |
| | | | 🏫 | 1 | 0 | 99 | 0 | 3 | 0 | 70 | 27 | 0 | 0 | 98 | 2 | 0 | 12 | 51 | 0 | 37 | 14 | 7 | 48 | 0 | 31 |
| | | | 💔 | 5 | 10 | 85 | 0 | 3 | 4 | 65 | 28 | 4 | 2 | 85 | 9 | 0 | 7 | 18 | 30 | 45 | 0 | 6 | 42 | 12 | 40 |
| | | | 🥋 | 5 | 2 | 93 | 0 | 2 | 7 | 80 | 11 | 0 | 2 | 95 | 3 | 0 | 4 | 59 | 1 | 36 | 0 | 0 | 11 | 2 | 87 |
| | 🫑 KSP | ➡️ | 🫑 | 5.3 | 68 | 25.3 | 1.3 | 10.7 | 72.7 | 4.7 | 12 | 31.3 | 60 | 7.3 | 1.3 | 92.7 | 6.7 | 0.7 | 0 | 0 | 45.3 | 19.3 | 35.3 | 0 | 0 |
| | | | 🍋 | 10.7 | 71.3 | 12 | 6 | 17.3 | 43.3 | 32 | 7.3 | 28.7 | 49.3 | 18.7 | 3.3 | 68.7 | 24 | 5.3 | 0.7 | 1.3 | 36.7 | 17.3 | 40.7 | 0 | 5.3 |
| | | | 🏛️ | 9.3 | 57.3 | 33.3 | 0 | 13.3 | 64 | 4.7 | 18 | 21.3 | 64 | 14.7 | 0 | 92 | 4 | 4 | 0 | 0 | 49.3 | 16 | 33.3 | 0 | 1.3 |
| | | | 🎉 | 11.3 | 52 | 36.7 | 0 | 16.7 | 68.7 | 6.7 | 8 | 26.7 | 50.7 | 22 | 0.7 | 92 | 6.7 | 1.3 | 0 | 0 | 46.7 | 10.7 | 42.7 | 0 | 0 |
| | | 🔄 | 🫑 | 14.7 | 18.7 | 54 | 12.7 | 26 | 54 | 14.7 | 5.3 | 14 | 42.7 | 36 | 7.3 | 88 | 0.7 | 7.3 | 0 | 4 | 53.3 | 11.3 | 35.3 | 0 | 0 |
| | | | 🍋 | 19.3 | 8.7 | 68.7 | 3.3 | 22 | 56.7 | 17.3 | 4 | 16.7 | 32 | 49.3 | 2 | 68.7 | 6.7 | 10.7 | 0 | 14 | 45.3 | 11.3 | 36.7 | 0.7 | 6 |
| | | | 🏛️ | 11.3 | 23.3 | 64.7 | 0.7 | 20.7 | 35.3 | 29.3 | 14.7 | 10 | 32.7 | 55.3 | 2 | 83.3 | 4.7 | 2.7 | 0 | 9.3 | 56 | 8 | 35.3 | 0 | 0.7 |
| | | | 🎉 | 8.7 | 28.7 | 54.7 | 8 | 26.7 | 32.7 | 26 | 14.7 | 13.3 | 33.3 | 45.3 | 8 | 87.3 | 0.7 | 10 | 0 | 2 | 36 | 14.7 | 49.3 | 0 | 0 |
| | ✈️ TSP | ➡️ | ✈️ | 8 | 92 | 0 | 0 | 12 | 62.7 | 1.3 | 24 | 6 | 94 | 0 | 0 | 1.3 | 2.7 | 0 | 0 | 96 | 13.3 | 36 | 8 | 6.7 | 36 |
| | | | 📋 | 5.3 | 94.7 | 0 | 0 | 7.3 | 72.7 | 0 | 20 | 6.7 | 93.3 | 0 | 0 | 0 | 2.7 | 0 | 0 | 97.3 | 8 | 18 | 2.7 | 9.3 | 62 |
| | | | 📅 | 3.3 | 96.7 | 0 | 0 | 9.3 | 80.7 | 2 | 8 | 4.7 | 94.7 | 0 | 0.7 | 0 | 4.7 | 2 | 2.7 | 90.7 | 3.3 | 20 | 0 | 0.7 | 76 |
| | | | 🪑 | 5.3 | 94.7 | 0 | 0 | 6 | 90.7 | 0 | 3.3 | 4.7 | 95.3 | 0 | 0 | 0.7 | 7.3 | 0.7 | 0 | 91.3 | 0 | 4.7 | 0 | 0 | 95.3 |
| | | 🔄 | ✈️ | 1.3 | 98.7 | 0 | 0 | 6.7 | 74 | 0.7 | 18.7 | 3.3 | 96 | 0 | 0.7 | 0.7 | 5.3 | 0 | 0.7 | 93.3 | 6 | 14 | 8 | 8.7 | 63.3 |
| | | | 📋 | 5.3 | 94.7 | 0 | 0 | 5.3 | 75.3 | 0.7 | 18.7 | 8 | 90.7 | 0 | 1.3 | 0 | 0 | 0.7 | 0.7 | 98.7 | 2.7 | 22 | 5.3 | 7.3 | 62.7 |
| | | | 📅 | 5.3 | 94.7 | 0 | 0 | 9.3 | 80.7 | 2.7 | 7.3 | 4.7 | 95.3 | 0 | 0 | 0 | 4 | 1.3 | 0 | 94.7 | 14 | 40 | 1.3 | 0.7 | 44 |
| | | | 🪑 | 7.3 | 92.7 | 0 | 0 | 6 | 90 | 0.7 | 3.3 | 4.7 | 95.3 | 0 | 0 | 0.7 | 2.7 | 1.3 | 0.7 | 94.7 | 0 | 0.7 | 0 | 0 | 99.3 |

Table 12: Full results for Qwen3-32B in non-thinking mode on both EHOP-RANDOM and EHOP-HARD, with formatting matching that of Table 10.

**ILP LP.** The ILP LP prompting strategy is very similar to ILP Python, with the exception that the LLM is asked to express the ILP program in the LP file format instead of as a Python program. We use the Gurobi solver (Gurobi Optimization LLC, 2024) to evaluate the code generated by the LLM, and we return the variable assignments generated by Gurobi in our follow-up message to the LLM. See our codebase for more details.

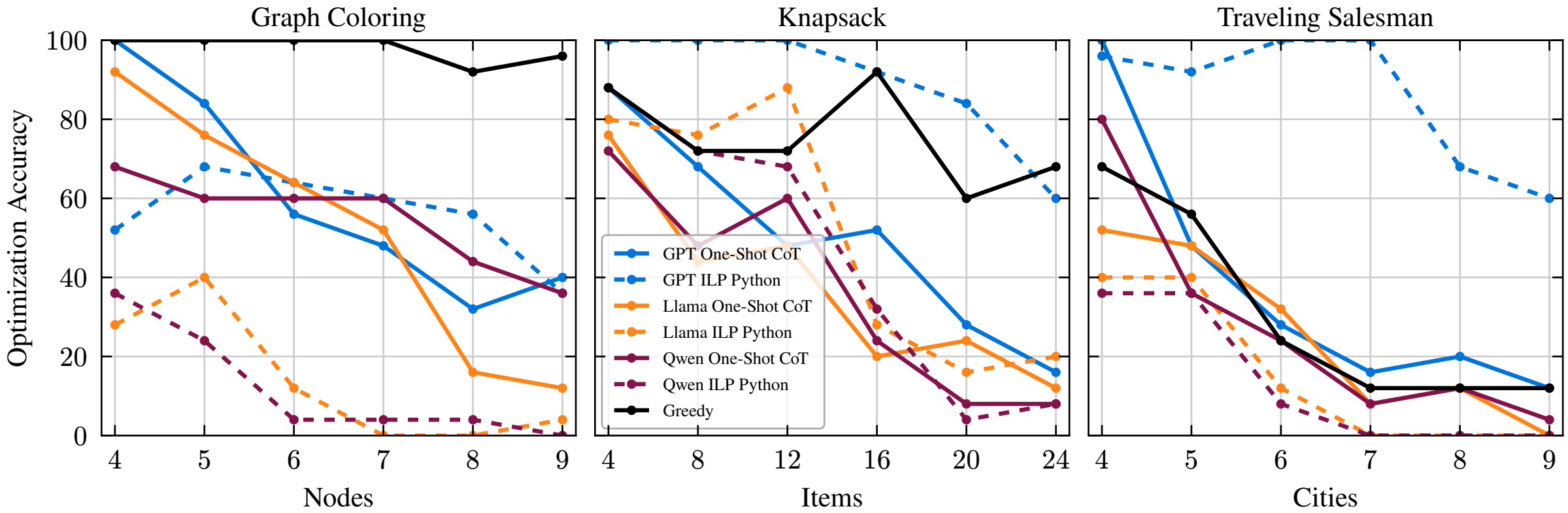

Figure 5: Optimization accuracy on the textbook variants in EHOP-RANDOM as a function of instance size.

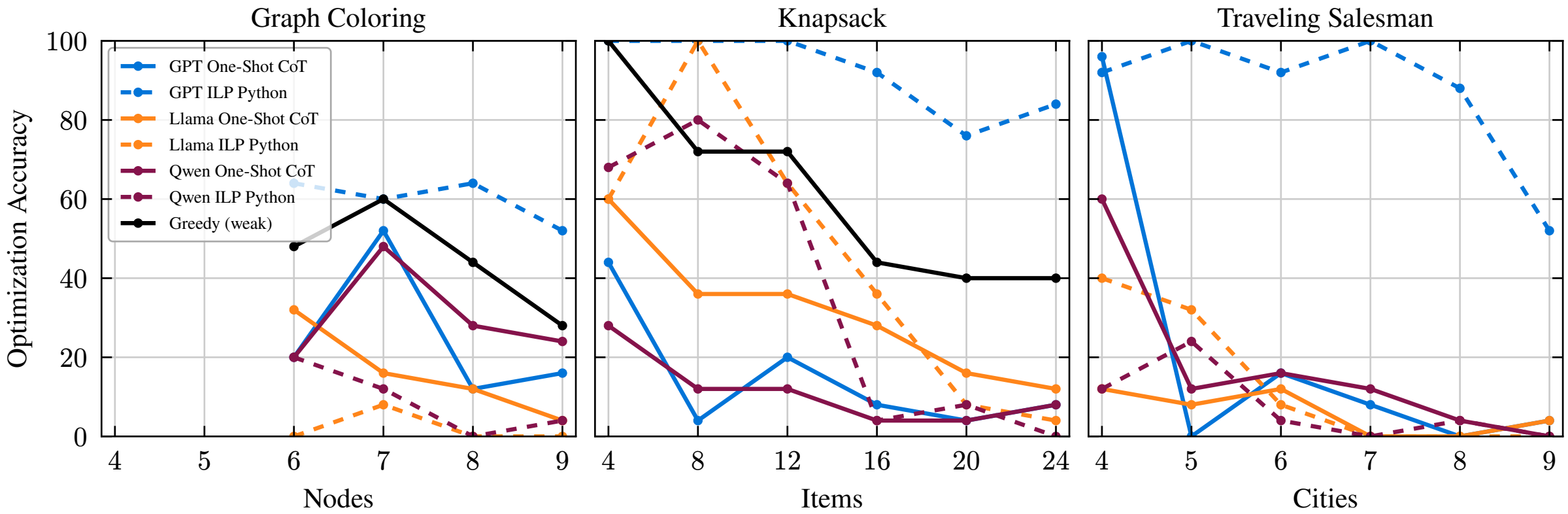

Figure 6: Optimization accuracy on the textbook variants in EHOP-HARD as a function of instance size.

## E Full Results

Table 10, Table 11, and Table 12 present full de-aggregated results from the experiments on GPT, Llama, and non-thinking Qwen, respectively; see Appendix H for reasoning model results. The tables break down results using the result categories discussed in Section 4.3.

## F Full Instance Size Effects

Figure 5 and Figure 6 present the effect of instance size on performance, as shown in Figure 3 and Figure 4, with results for Qwen3-32B in non-thinking mode included.

## G Results on OPRO

We present the detailed results of running OPRO on EHOP-RANDOM in Table 13. "Small" refers to the second-smallest instance sizes of each variant (e.g., five cities in TRAVELING SALESMAN), and "large" refers to the largest instance sizes in each variant (e.g. nine cities in TRAVELING SALESMAN). The numbers are for GPT-4o, the best-performing standard LLM in our evaluation.

| Problem | Variant | Small | Large |
|---|---|---|---|
| GCP | Textbook | 96 | 56 |
| GCP | Inverted | −48.0 | −56.0 |
| GCP | Costumed | −1.3 | −16.0 |
| KSP | Textbook | 96 | 24 |
| KSP | Inverted | +0.0 | +24.0 |
| KSP | Costumed | −8.0 | +6.7 |
| TSP | Textbook | 100 | 36 |
| TSP | Inverted | +0.0 | −36.0 |
| TSP | Costumed | +0.0 | −30.7 |

Table 13: Optimization accuracies for GPT-4o on EHOP-RANDOM using OPRO. Formatting as in Table 1.

## H Reasoning Model Results

Table 14 presents full de-aggregated results from the experiments on DeepSeek-R1 and thinking-mode Qwen.

DeepSeek-R1 engages in very long reasoning chains in which it frequently attempts to identify the base problem from which the instance is

| | | | | EHOP-RANDOM | | | | | | | | | EHOP-HARD | | | | | | | | |
|---|---|---|---|---|---|---|---|---|---|---|---|---|---|---|---|---|---|---|---|---|---|
| | | | | Zero-Shot | | | | ILP Python | | | | | Zero-Shot | | | | ILP Python | | | | |
| | | | | O | S | E | I | O | S | E | I | F | O | S | E | I | O | S | E | I | F |
| 🖌️ GCP | R1 | ➡️ | 🖌️ | 100 | 0 | 0 | 0 | 91.3 | 0 | 0.7 | 3.3 | 4.7 | 98 | 2 | 0 | 0 | 94 | 1 | 0 | 2 | 3 |
| | | | 🏨 | 97.3 | 0.7 | 0 | 2 | 97.3 | 0 | 0 | 1.3 | 1.3 | 97 | 1 | 0 | 2 | 99 | 0 | 0 | 0 | 1 |
| | | | 💔 | 98.7 | 0 | 0 | 1.3 | 97.3 | 0 | 0 | 2 | 0.7 | 95 | 3 | 0 | 2 | 96 | 0 | 0 | 1 | 3 |
| | | | 🥋 | 95.3 | 0 | 0 | 4.7 | 96.7 | 0 | 0 | 0 | 3.3 | 90 | 5 | 0 | 5 | 97 | 0 | 0 | 0 | 3 |
| | | 🔄 | 🖌️ | 38 | 0 | 49.3 | 12.7 | 51.3 | 0 | 44.7 | 3.3 | 0.7 | 23 | 0 | 69 | 8 | 38 | 1 | 55 | 4 | 2 |
| | | | 🏨 | 86.7 | 0 | 0.7 | 12.7 | 98.7 | 0 | 0 | 0.7 | 0.7 | 94 | 0 | 0 | 6 | 97 | 0 | 0 | 2 | 1 |
| | | | 💔 | 79.3 | 0 | 0 | 20.7 | 95.3 | 0 | 0 | 3.3 | 1.3 | 91 | 0 | 0 | 9 | 99 | 0 | 0 | 1 | 0 |
| | | | 🥋 | 77.3 | 0.7 | 0 | 22 | 98.7 | 0 | 0.7 | 0 | 0.7 | 80 | 2 | 0 | 18 | 98 | 0 | 0 | 2 | 0 |
| | Qwen | ➡️ | 🖌️ | 77.3 | 1.3 | 0 | 21.3 | 76 | 1.3 | 4.7 | 5.3 | 12.7 | 62 | 2 | 0 | 36 | 72 | 4 | 8 | 9 | 7 |
| | | | 🏨 | 80 | 2.7 | 0.7 | 16.7 | 80 | 1.3 | 5.3 | 2.7 | 10.7 | 69 | 7 | 0 | 24 | 81 | 4 | 5 | 3 | 7 |
| | | | 💔 | 74.7 | 3.3 | 0.7 | 21.3 | 72 | 2.7 | 4.7 | 5.3 | 15.3 | 77 | 2 | 0 | 21 | 67 | 1 | 9 | 5 | 18 |
| | | | 🥋 | 78 | 2 | 0.7 | 19.3 | 84.7 | 0.7 | 3.3 | 2.7 | 8.7 | 74 | 4 | 0 | 22 | 87 | 1 | 0 | 3 | 9 |
| | | 🔄 | 🖌️ | 34.7 | 0 | 0 | 65.3 | 75.3 | 2 | 5.3 | 1.3 | 16 | 12 | 0 | 1 | 87 | 73 | 4 | 6 | 8 | 9 |
| | | | 🏨 | 42.7 | 0.7 | 2.7 | 54 | 80 | 1.3 | 1.3 | 5.3 | 12 | 35 | 0 | 4 | 61 | 64 | 3 | 7 | 5 | 21 |
| | | | 💔 | 41.3 | 0 | 0 | 58.7 | 76 | 0 | 4 | 3.3 | 16.7 | 24 | 0 | 1 | 75 | 66 | 1 | 11 | 8 | 14 |
| | | | 🥋 | 37.3 | 0 | 4 | 58.7 | 76.7 | 0 | 2 | 4.7 | 16.7 | 17 | 1 | 5 | 77 | 78 | 1 | 7 | 6 | 8 |
| 🎒 KSP | R1 | ➡️ | 🎒 | 62.7 | 37.3 | 0 | 0 | 98 | 2 | 0 | 0 | 0 | 48.7 | 51.3 | 0 | 0 | 97.3 | 2.7 | 0 | 0 | 0 |
| | | | 🍋 | 52 | 48 | 0 | 0 | 98.7 | 0.7 | 0 | 0.7 | 0 | 40.7 | 59.3 | 0 | 0 | 98 | 0 | 1.3 | 0 | 0.7 |
| | | | 🏛️ | 54.7 | 45.3 | 0 | 0 | 99.3 | 0.7 | 0 | 0 | 0 | 52.7 | 47.3 | 0 | 0 | 98.7 | 1.3 | 0 | 0 | 0 |
| | | | 🎉 | 80.7 | 19.3 | 0 | 0 | 100 | 0 | 0 | 0 | 0 | 68 | 32 | 0 | 0 | 100 | 0 | 0 | 0 | 0 |
| | | 🔄 | 🎒 | 67.3 | 0 | 32.7 | 0 | 98.7 | 0 | 0.7 | 0 | 0.7 | 62.7 | 0 | 37.3 | 0 | 98 | 0 | 0.7 | 0 | 1.3 |
| | | | 🍋 | 59.3 | 0.7 | 40 | 0 | 99.3 | 0 | 0 | 0.7 | 0 | 60.7 | 0.7 | 38.7 | 0 | 99.3 | 0 | 0 | 0 | 0.7 |
| | | | 🏛️ | 76 | 1.3 | 22.7 | 0 | 100 | 0 | 0 | 0 | 0 | 76.7 | 2 | 21.3 | 0 | 98.7 | 0 | 1.3 | 0 | 0 |
| | | | 🎉 | 67.3 | 1.3 | 31.3 | 0 | 100 | 0 | 0 | 0 | 0 | 66 | 0 | 34 | 0 | 100 | 0 | 0 | 0 | 0 |
| | Qwen | ➡️ | 🎒 | 28.7 | 0 | 1.3 | 70 | 86.7 | 0.7 | 0.7 | 0.7 | 11.3 | 23.3 | 0 | 0 | 76.7 | 90 | 0 | 0 | 0.7 | 9.3 |
| | | | 🍋 | 25.3 | 0 | 0 | 74.7 | 77.3 | 1.3 | 0.7 | 0.7 | 20 | 20.7 | 0 | 0 | 79.3 | 78 | 3.3 | 0 | 0.7 | 18 |
| | | | 🏛️ | 28 | 0.7 | 0 | 71.3 | 81.3 | 2 | 0.7 | 0 | 16 | 25.3 | 0 | 0 | 74.7 | 84.7 | 2 | 0.7 | 0 | 12.7 |
| | | | 🎉 | 28.7 | 0 | 0 | 71.3 | 77.3 | 2 | 2 | 0 | 18.7 | 20 | 0 | 0 | 80 | 86.7 | 0.7 | 1.3 | 0 | 11.3 |
| | | 🔄 | 🎒 | 28 | 0 | 0 | 72 | 81.3 | 1.3 | 0.7 | 0 | 16.7 | 21.3 | 0.7 | 0 | 78 | 82.7 | 1.3 | 0 | 0 | 16 |
| | | | 🍋 | 24 | 0.7 | 0 | 75.3 | 68.7 | 2 | 2 | 0 | 27.3 | 20.7 | 0.7 | 0 | 78.7 | 80 | 4 | 1.3 | 0 | 14.7 |
| | | | 🏛️ | 28.7 | 2 | 0 | 69.3 | 67.3 | 3.3 | 2.7 | 2.7 | 24 | 23.3 | 0 | 0 | 76.7 | 76 | 2.7 | 4.7 | 0.7 | 16 |
| | | | 🎉 | 26 | 0 | 0 | 74 | 80.7 | 0.7 | 2 | 0 | 16.7 | 20.7 | 0 | 0 | 79.3 | 83.3 | 2.7 | 1.3 | 0 | 12.7 |
| ✈️ TSP | R1 | ➡️ | ✈️ | 34.7 | 0 | 0 | 65.3 | 82 | 0 | 0 | 14.7 | 3.3 | 32 | 0 | 0 | 68 | 72.7 | 0 | 0 | 24 | 3.3 |
| | | | 📋 | 26.7 | 0 | 0 | 73.3 | 92.7 | 0 | 0 | 4 | 3.3 | 22.7 | 0 | 0 | 77.3 | 88.7 | 0 | 0 | 9.3 | 2 |
| | | | 📅 | 20 | 0 | 0 | 80 | 86.7 | 0 | 0 | 6.7 | 6.7 | 18.7 | 0 | 0 | 81.3 | 89.3 | 0 | 0 | 6 | 4.7 |
| | | | 🪑 | 21.3 | 0 | 0 | 78.7 | 50 | 0 | 0 | 24.7 | 25.3 | 22.7 | 0 | 0 | 77.3 | 52.7 | 0 | 0 | 26 | 21.3 |
| | | 🔄 | ✈️ | 28 | 0 | 0 | 72 | 71.3 | 0 | 0 | 22.7 | 6 | 31.3 | 0 | 0 | 68.7 | 81.3 | 1.3 | 0 | 14 | 3.3 |
| | | | 📋 | 25.3 | 0 | 0 | 74.7 | 86.7 | 0.7 | 0 | 7.3 | 5.3 | 26 | 0 | 0 | 74 | 92.7 | 0 | 0 | 4 | 3.3 |
| | | | 📅 | 29.3 | 0 | 0 | 70.7 | 88 | 0 | 0 | 6.7 | 5.3 | 24 | 0.7 | 0 | 75.3 | 84.7 | 0 | 0 | 6 | 9.3 |
| | | | 🪑 | 32 | 0.7 | 0 | 67.3 | 50 | 0 | 0 | 24.7 | 25.3 | 30 | 1.3 | 0 | 68.7 | 54 | 0 | 0 | 25.3 | 20.7 |
| | Qwen | ➡️ | ✈️ | 14.7 | 0 | 0 | 85.3 | 56 | 2 | 0.7 | 4.7 | 36.7 | 14 | 0 | 0 | 86 | 50.7 | 3.3 | 0 | 6 | 40 |
| | | | 📋 | 14.7 | 0 | 0 | 85.3 | 42.7 | 5.3 | 0 | 2.7 | 49.3 | 12.7 | 0 | 0 | 87.3 | 43.3 | 4.7 | 0 | 3.3 | 48.7 |
| | | | 📅 | 15.3 | 0 | 0 | 84.7 | 43.3 | 10.7 | 0.7 | 2 | 43.3 | 14 | 0 | 0 | 86 | 48.7 | 16.7 | 0 | 5.3 | 29.3 |
| | | | 🪑 | 11.3 | 0 | 0 | 88.7 | 45.3 | 3.3 | 0 | 2 | 49.3 | 8.7 | 0 | 0 | 91.3 | 46 | 8.7 | 0 | 0 | 45.3 |
| | | 🔄 | ✈️ | 16.7 | 0 | 0 | 83.3 | 56 | 6.7 | 0 | 0.7 | 36.7 | 17.3 | 0 | 0 | 82.7 | 55.3 | 4 | 0 | 2.7 | 38 |
| | | | 📋 | 14.7 | 0 | 0 | 85.3 | 48.7 | 10.7 | 0 | 2 | 38.7 | 13.3 | 0 | 0 | 86.7 | 40.7 | 10.7 | 0 | 2.7 | 46 |
| | | | 📅 | 14.7 | 0 | 0 | 85.3 | 45.3 | 12.7 | 0 | 2.7 | 39.3 | 11.3 | 0 | 0 | 88.7 | 46 | 18.7 | 0 | 3.3 | 32 |
| | | | 🪑 | 12 | 0 | 0 | 88 | 46.7 | 2.7 | 0 | 2 | 48.7 | 10 | 0 | 0 | 90 | 40.7 | 4 | 0 | 1.3 | 54 |

Table 14: Full results for DeepSeek-R1 and thinking-mode Qwen3 on both EHOP-RANDOM and EHOP-HARD (see topmost column headers), with formatting matching that of Table 10.

derived. It frequently does this successfully, and it then uses specific knowledge about the base problem to solve the instance. In order to quantify this behavior, we searched the DeepSeek reason-ing logs for occurrences of the name of the base problem (e.g., "knapsack"). The proportion of instances in which the base problem name or something closely related (e.g., "chromatic number" for graph coloring) appeared is shown in Table 15. Note that for inverted problems, a mention of the base problem can mean that DeepSeek generated a thought of the form "this almost looks like Knapsack, but …"

| Problem | Prompting Strategy | GPT | | | Llama | | |
|---|---|---|---|---|---|---|---|
| | | Optimal | Suboptimal | Erroneous | Optimal | Suboptimal | Erroneous |
| GCP | One-Shot | 30.9 | 7.3 | 61.8 | 1.3 | 0.9 | 97.8 |
| | Zero-Shot CoT | 40.7 | 12.3 | 46.9 | 28.4 | 18.0 | 53.6 |
| | One-Shot CoT | 28.6 | 10.5 | 60.9 | 17.1 | 18.5 | 64.4 |
| | ILP LP | 7.1 | 10.6 | 82.3 | 2.3 | 19.5 | 78.2 |
| | ILP Python | 46.5 | 12.0 | 41.5 | 18.5 | 12.3 | 69.2 |
| KSP | One-Shot | 0.4 | 67.2 | 32.3 | 0.9 | 73.6 | 25.5 |
| | Zero-Shot CoT | 0.0 | 18.1 | 81.9 | 0.5 | 34.9 | 64.6 |
| | One-Shot CoT | 0.6 | 26.0 | 73.4 | 0.0 | 45.2 | 54.8 |
| | ILP LP | 0.0 | 60.0 | 40.0 | 0.0 | 90.5 | 9.5 |
| | ILP Python | 0.0 | 63.9 | 36.1 | 0.0 | 85.1 | 14.9 |
| TSP | One-Shot | 1.1 | 98.9 | 0.0 | 0.7 | 99.3 | 0.0 |
| | Zero-Shot CoT | 0.5 | 97.6 | 1.9 | 0.5 | 99.0 | 0.5 |
| | One-Shot CoT | 0.9 | 99.1 | 0.0 | 1.1 | 98.9 | 0.0 |
| | ILP LP | 0.0 | 95.2 | 4.8 | 0.0 | 100.0 | 0.0 |
| | ILP Python | 0.0 | 89.7 | 10.3 | 1.2 | 65.0 | 33.8 |

Table 16: The result types of suboptimal/erroneous Inverted instances, when interpreted as Textbook.

| Problem | Variant | Zero-Shot | | ILP Python | |
|---|---|---|---|---|---|
| | | R1 | Qwen | R1 | Qwen |
| GCP | Textbook | 93 | 58 | 100 | 100 |
| | Inverted | 100 | 26 | 100 | 98 |
| | Costumed | 97 | 78 | 99 | 100 |
| KSP | Textbook | 86 | 10 | 86 | 100 |
| | Inverted | 56 | 14 | 17 | 60 |
| | Costumed | 79 | 9 | 51 | 36 |
| TSP | Textbook | 97 | 14 | 100 | 87 |
| | Inverted | 51 | 16 | 98 | 89 |
| | Costumed | 46 | 8 | 70 | 56 |

Table 15: Percentage of instances of EHOP in which R1 and thinking-mode Qwen3 recognize the base problem.

## I Inverted interpreted as standard

As we discussed in Section 5.5, one potential error pattern on Inverted instances is that the LLM misses the distinction between Inverted and Textbook and accidentally solves Textbook instead. If this error pattern is frequent, we should be seeing a lot of Inverted instances with a solution that was suboptimal or erroneous for Inverted, but optimal for the Textbook instance with the same parameters. For instance, a solution to Inverted GRAPH COLORING has to give non-adjacent nodes different colors. A solution in which the *adjacent* nodes all have different colors will be erroneous for Inverted GRAPH COLORING, but may be optimal for Textbook GRAPH COLORING.

Table 16 shows the results of an evaluation to measure this effect. It shows e.g. that out of all the instances of Inverted GRAPH COLORING on which GPT-4o with the One-Shot strategy gave a suboptimal or erroneous solution, 30.9% of those instances were an optimal solution to Textbook GRAPH COLORING. In general, most non-optimal solutions remain non-optimal for Textbook, indicating that this error pattern does not actually apply.

## J Mentioning the base problem in the prompt

Table 17 shows the optimization accuracies for GPT-4o on EHOP-RANDOM. The "hinted" columns indicate that the prompt mentions the base problem; the others are as in Table 1.

Note that unlike in all other tables, "Costumed*" in Table 17 is not the average over all three costumes, but the results on a single costume. For GRAPH COLORING, this is Student Groups; for KNAPSACK, it is Lemonade Stand; and for TRAVELING SALESMAN, it is UN Seating.

| Problem | Variant | One-Shot CoT | One-Shot CoT (hinted) | ILP Python | ILP Python (hinted) |
|---|---|---|---|---|---|
| GCP | Textbook | 60.0 | 66 | 56 | 48 |
| | Inverted | −59.3 | −66.0 | −41.3 | −45.3 |
| | Costumed* | −4.7 | −3.3 | −43.8 | +14.7 |
| KSP | Textbook | 50 | 34 | 89.3 | 88.7 |
| | Inverted | −4.7 | +8.7 | −0.6 | −4.7 |
| | Costumed* | −2.2 | +18.0 | −7.5 | −0.7 |
| TSP | Textbook | 37.3 | 36.7 | 86 | 89.3 |
| | Inverted | −9.3 | −8.0 | −10.7 | −11.3 |
| | Costumed* | −9.1 | −6.0 | −37.1 | −42.6 |

Table 17: Optimization accuracy for GPT-4o on EHOP-RANDOM with and without hinting what the base problem is. Formatting matches that of Table 1.

## K Knapsack Performance Anomalies

There is a noticeable irregularity in the results for the KNAPSACK domain in the EHOP-HARD set. This is the one group where inversion and costuming appear to help more than they hurt. We have not isolated a sole cause of this discrepancy, but we have identified several factors that we expect to have an effect on performance that could be different for KNAPSACK:

- **Frequency of mentions in training data:** If KNAPSACK is discussed more or less often in the LLMs' training data than GRAPH COLORING and TRAVELING SALESMAN, this could result in variability in the depth/complexity of the models' learned associations for the problem, potentially making them better equipped to solve KNAPSACK instances regardless of variant.
- **Quality of explanations in training data:** Similarly, the quality/depth of analysis presented in documents discussing the knapsack problem, as well as the consistency of this quality across documents, could influence the chains of thought generated by an LLM when solving KNAPSACK instances.
- **The nature of the problem:** KNAPSACK is distinct from GRAPH COLORING and TRAVELING SALESMAN in that it is a purely numeric problem, whereas the others involve graphs. It is possible that this makes KNAPSACK easier to solve in a text-only format (as well as more likely to appear in text-only analyses on the internet), both of which could contribute to a different pattern of performance for this base problem.
- **The degree to which costumes successfully disguise the base problem:** While all costumes were meant to be equally misleading, it is possible that the KNAPSACK costumes were more straightforward or were somehow clearer than for other base problems.
- **Inversion mechanics:** The precise way in which a base problem is inverted varies across all three problems: inverted GRAPH COLORING maintains the objective (minimize colors) while flipping the constraint (non-adjacent nodes must not match); inverted KNAPSACK flips the objective (minimize value) and the constraint (total weight cannot be below capacity); inverted TRAVELING SALESMAN flips the objective (maximize distance) and maintains the constraint (each node is visited once before returning to the start). It is possible that the unique combination of flipping both the objective and constraint of KNAPSACK lead to an inverted problem that is easier for LLMs to solve than the original.
- **Instance generation:** Given that all base problems vary in their structure, the methods we used for randomly generating instances also varied. It is thus possible that the methods for generating KNAPSACK instances produced a set of problems that somehow differed in their difficulty distribution relative to the other two problems.

These factors and more could all be potential sources of variation in performance across base problems and could thus explain the anomalous behavior on KNAPSACK. Future research efforts could investigate which factors are more likely to be the cause of the trends we observed.